\documentclass[journal]{IEEEtran}

\usepackage{booktabs}
\usepackage{cite}
\usepackage{hyperref}
\usepackage{xspace}
\usepackage{graphicx}
\usepackage{caption}
\usepackage{rotating}
\usepackage{multirow,bigdelim}
\usepackage{comment}
\usepackage{pifont}
\usepackage{xcolor}
\usepackage{wrapfig}

\newcommand*\rotvertical{\rotatebox{90}}
\newcommand{\myparagraph}[1]{\smallskip \noindent \textbf{#1.}}
\newcommand{\question}[1]{Q{#1}}
\newcommand{\quoteP}[1]{``\emph{#1}''}
\newcommand{\security}{security\xspace}
\newcommand{\safety}{safety\xspace}

\newcommand{\nonMLSecurity}{non-AML security\xspace}

\newcommand{\checkmark}{\ding{51}}
\newcommand{\classifier}{\ensuremath{f}\xspace}
\newcommand{\params}{\ensuremath{\omega}\xspace}

\newcommand{\trainx}{\ensuremath{X}\xspace}
\newcommand{\trainy}{\ensuremath{Y}\xspace}
\newcommand{\testx}{\ensuremath{X_t}\xspace}
\newcommand{\testy}{\ensuremath{Y_t}\xspace}

\begin{document}

\title{Machine Learning Security in Industry: \\ A Quantitative Survey}

%
%
%

\author{Kathrin Grosse, 
        Lukas Bieringer, 
        Tarek R. Besold, 
        Battista Biggio,~\IEEEmembership{Senior~Member,~IEEE,}
        Katharina Krombholz
\thanks{First two authors contributed equally. K. Grosse is with the EPFL; L.~Bieringer with QuantPi; T.R. Besold with Eindhoven University of Technology; K. Krombholz with CISPA Helmholtz Center of Information Security; and B. Biggio with the University of Cagliari. 
}}

%
%

\markboth{Grosse \MakeLowercase{\textit{et al.}}: Machine Learning Security in Industry}%
{Grosse \MakeLowercase{\textit{et al.}}: Machine Learning Security in Industry}
%



\maketitle

\begin{abstract}
Despite the large body of academic work on machine learning security, little is known about the occurrence of attacks on machine learning systems in the wild. In this paper, we report on a quantitative study with 139 industrial practitioners. We analyze attack occurrence and concern and evaluate statistical hypotheses on factors influencing threat perception and exposure. Our results shed light on real-world attacks on deployed machine learning. On the organizational level, while we find no predictors for threat exposure in our sample, the amount of implement defenses depends on exposure to threats or expected likelihood to become a target. We also provide a detailed analysis of practitioners' replies on the relevance of individual machine learning attacks, unveiling complex concerns like unreliable decision making, business information leakage, and bias introduction into models. Finally, we find that on the individual level, prior knowledge about machine learning security influences threat perception. Our work paves the way for more research about adversarial machine learning in practice, but yields also insights for regulation and auditing.
\end{abstract}


%
\IEEEpeerreviewmaketitle

\section{Introduction}
A large body of academic work focuses on machine learning security or adversarial machine learning (AML)~\cite{barreno2006can,biggio2018wild,chen2017targeted,cina2022wild,Dalvi:2004:AC:1014052.1014066,gu_badnets_2017,ji2017backdoor,oh2019towards,papernot2016transferability,DBLP:journals/corr/SzegedyZSBEGF13,DBLP:conf/uss/TramerZJRR16}. These works investigate how machine learning (ML) can be circumvented and exploited by an attacker. 
For example, an attacker can tamper with the training data, yielding a model inferior in performance or that is sensitive to attacker specified, small parts of the input~\cite{cina2022wild}.
Alternatively, the attacker slightly alters test data to change the output of an ML model~\cite{Dalvi:2004:AC:1014052.1014066,DBLP:journals/corr/SzegedyZSBEGF13}. In addition, an ML model may leak the used training data~\cite{2016arXiv161005820S} or can easily be copied when freely exposed~\cite{DBLP:conf/uss/TramerZJRR16}. 

Many of the settings studied in ML security can be criticized for being rather artificial. However, already these settings are hard to solve~\cite{cina2022wild,tramer2020adaptive}. One possible cause is that even though the current usage of ML in security and threat modelling have been criticised\cite{sommer2010outside,gilmer2018motivating}, there is little work on ML security in the real world. 
In the first work in this direction, by Kumar et al.~\cite{kumar2020adversarial} investigated which AML threats are feared in practice by interviewing 28 organizations whose largest concern were poisoning attacks.
 Mirsky et al.~\cite{mirsky2021threat} reported that the 22 interviewed organizations perceived 24 of 33 offensive AI techniques as a significant threat.
 Moreover, Bieringer et al.~\cite{bieringer2021mental} found evidence for rudimentary attacks on AI in the wild in their 15 qualitative interviews with ML practitioners. 
 Boenisch et al.~\cite{boenisch2021never} found that in their 83 participants, security and ML security awareness of ML practitioners was overall low. 
In contrast to these previous works, our sample with 139 participants is larger and more diverse. Our focus only on ML security (not considering privacy, or offensive AI) allows us further to study threat concern in depth, and to run statistical tests on our participant's replies. Finally, we are the first to publish an estimate about ML security incidents from the real world but not covered in media.

More specifically, to shed light on the state of AML in practice and the factors influencing organizations' approach to ML security, we conduct a quantitative survey among ML practitioners. 
Inspired by prior work~\cite{mirsky2021threat}, we investigate which threats are dreaded and why. Furthermore, 
we control for variables like application area~\cite{boenisch2021never}, 
how long ML has been used in production,
data type, and prior knowledge~\cite{bieringer2021mental}. 
All these questions and variables form part of our anonymous questionnaire for ML-practitioners. Our 139 participants help us to shed light on the following topics:

\myparagraph{AML in practice} We find that there are occurrences of AML attacks, more specifically evasion and poisoning, in practice. However, non-ML security threats (e.g., access control, botnets, resource theft, etc.) are also prevalent and (still) seem to pose a larger concern, together with organizational challenges, privacy and benign ML challenges.

\myparagraph{AML within organizations} We find that exposure to AML threats is not related to organization size, time that ML is used in production, or organization area. Yet, some of these factors, together with exposure, influence strongly how many mitigations an organization has in place. Furthermore, we investigate why practitioners deem an AML attack as relevant, and find a complex array of reasons, including business or financial, even ethical concerns. When an attack is judged as irrelevant, this is often a consequence of an application or deployment setting that makes the attack infeasible. 

\myparagraph{AML for practitioners} We find that self reported prior knowledge, in particular in AML, increases the concern reported for individual attacks. Concerning gender, we find that two of the five attacks tested are rated significantly less important by women, the remaining threats are rated similarly.

Our results open the avenue of more in depth studies that encompass application and deployment when studying vulnerability. Our results also shed light on the relationship between knowledge and threat perception. Our insights are furthermore valuable when regulating and auditing ML systems, as we a) analyze the underlying reasons for relevance or irrelevance of specific AML attacks and b) show that \security and \safety are conflated by our participants. The latter refer to the difference of benign system failures (safety) and attacker induced failures (security), as used in system analyses~\cite{muller2007personal}. We finally deduce that ML security incidents in practice are not as common as for example \nonMLSecurity, but that monitoring ML security might be beneficial.  

\section{Background}\label{sec:AML}
Before we describe our questionnaire and the methodology of our study, we would like to provide background on ML security or AML, as we confronted participants with the most relevant attacks discussed in AML theory~\cite{biggio2018wild} and practice~\cite{kumar2020adversarial}. 
More concretely, we focus on the six most relevant attacks in the industrial ranking by Kumar et al.~\cite{kumar2020adversarial}. 
We now give a rough overview of these attacks, and refer the interested reader to Bieringer et al.~\cite{bieringer2021mental} or recent surveys~\cite{biggio2018wild,cina2022wild}. To ease understanding of the below attacks, we first define ML formally. In ML, a by \params parametrized function \classifier is optimized during training to fit the training data \trainx, \trainy. After training, $\classifier (\trainx, \params) = \trainy$, and we expect the classifier to generalize to unseen test data, e.g. $\classifier (\testx ,\params) \approx \testy$. We further write $\trainx^*$ when an attacker has altered, or perturbed, the benign data \trainx.

\myparagraph{Poisoning} Poisoning affects, via the training data, the classifier at test time to reduce the overall performance or accuracy. To this end, the attacker manipulates samples $\trainx^*$~\cite{rubinstein2009antidote}, labels $\trainy^*$~\cite{biggio2011support} or both before training. Poisoning defenses are, compared to most attacks, well understood in terms of trade-offs, for example when it comes to attack strength and detectability~\cite{cina2022wild}. We also investigate backdoors, a variant of poisoning which affects, via the training data $\{\trainx^*,\trainy^*\}$, a specific subset of the test data $\{\testx, \testy \}$~\cite{chen2017targeted}. 

\myparagraph{Evasion} Evasion or adversarial example affect, via the perturbed test data $\testx^*$, the classifier \classifier and forces it to output either a predefined or a wrong output for an input, hence $\classifier (\testx^* , \params) \neq \testy$. To this end, the attacker changes the test input of a trained classifier carefully based on the 
classifier ~\cite{Dalvi:2004:AC:1014052.1014066,DBLP:journals/corr/SzegedyZSBEGF13}. Most evasion defenses are caught in an arms-race~\cite{tramer2020adaptive}.  

\myparagraph{Membership inference} 
While the previous attacks harmed performance, this attack harms the privacy of the training data \trainx. More specifically, the attacker queries the model at test time to deduce whether a point was used in training, e.g. $x \in \trainx $~\cite{2016arXiv161005820S}. Against these attacks, defenses have been proposed~\cite{nasr2018machine}, but they are not as well understood as poisoning or evasion, for example.

\myparagraph{Model stealing} 
Finally, in model stealing, the attacker harms the intellectual property of the model owner by copying the model \classifier without consent. 
To perform this attack, the attacker queries the model with the goal to use the obtained data to steal the model~\cite{DBLP:conf/uss/TramerZJRR16}. For this attack as well, defenses have been introduced~\cite{juuti2019prada}, nut no consensus has been reached so far on a standard defense.

Kumar et al.~\cite{kumar2020adversarial} distinguish model stealing and model extraction, a distinction that we avoided to decrease the risk of confusion for our participants.  
Beyond evidence by Bieringer et al.~\cite{bieringer2021mental},  Lin and Biggio~\cite{lin2021adversarial} and Kumar et al.~\cite{kumar2020adversarial}, few works have studied the relevance of these attacks in practice.

\section{Methodology}\label{sec:methodology}
Given this limited knowledge about AML in practice, we opted to design an questionnaire study that allows to reach many participants, yet still allows to process open text answers concerning for example the relevance of attacks. In this section, we describe our initial estimates on participants and how we designed our questionnaire. 

\myparagraph{Power analysis}
Before we started designing our questionnaire, we needed to know how many participants we required: few participants would allow a longer questionnaire, more required a shorter.
Many of our research hypotheses were testable with ordinal regression, as we planed to investigate how ordinal factors like organization size or organization maturity influence exposure to threats, for example. 
To use this regression with a power of $0.8$, a medium effect and a significance level of $0.05$, we would need, depending on the number of predictors, between 67 (2 predictors) and 97 (6 predictors) participants. Green's rule of thumb provides similar results. 
An alternative for two nominal variables is the Mann-Whitney-U test. 
This test assume as $H_0$-hypothesis that the two samples follow the same distribution, and is sufficiently powerful already for small sample sizes of 20~\cite{nachar2008mann}.

\subsection{Questionnaire design}
Given the required sample size of roughly 100, we opted for an anonymous survey with in total 32 questions. The questionnaire contained open-ended questions, multiple choice questions, checkboxes, and relevance rankings based on a Likert scale. For checkboxes and multiple choice questions, the order of all replies was randomized to avoid order bias~\cite{ferber1952order}. Questions, descriptions as well as the wording of answer options for multiple choice questions were based on prior research. In the following, we detail references used for the questionnaire along it's three parts, (1) AML in practice, (2) organizational background of participants, and (3) individual background of participants.
The complete questionnaire can be found in App.~\ref{app:questionnaire}. 

We decided not to pay our participants to avoid money-driven participation.
Furthermore, To not restrict our participants to ML specifically, we used the term artificial intelligence (AI) throughout the questionnaire, although our analysis focuses on ML security.

\myparagraph{Questionnaire--AML in practice} The first part of our questionnaire addressed security within participants’ AI workflows, products or systems. This included an open-ended question about the most pressing security challenges in participants' daily work and an indication on whether they had already experienced a circumvention of AI based workflows, products or systems~\cite{bieringer2021mental}. In addition, we asked participants to estimate their risk to become a victim of attacks on their AI based systems during the next 12 months and to provide information about their organization's approaches towards AI security~\cite{huaman2021large}. Choice and concrete wording of measures for AI security have been based on prior research~\cite{mandia2001incident}, dictionaries,\footnote{\url{https://www.merriam-webster.com}} recent regulatory approaches,\footnote{e.g. Artificial Intelligence Act by European Commission} and auditing frameworks.\footnote{e.g. AI Cloud Service Compliance Criteria Catalogue by German Federal Office for Information Security, AI Auditing Catalogue by Fraunhofer IAIS} Following Huaman et al.~\cite{huaman2021large}, we further inquired the relevance of the previously discussed relevant ML attacks~\cite{biggio2018wild,kumar2020adversarial}. For each attack and a made-up sanity threat, participants were shown a description (see App.~\ref{app:questionnaire}) to assess the attack's relevance.

\myparagraph{Questionnaire--organizational background} We also queried information on participants' organizations and their AI practices. This included basic information like the number of employees~\cite{european2003commission} or industry area~\cite{kumar2020adversarial}. With regards to AI practices, we queried information with regards to organizations' primary data analysis type (supervised/unsupervised learning, reinforcement learning) and input data (images, program code, etc.) 
and labels (real valued, discrete, etc). In addition, we asked for the status of AI projects in the organization, for example \quoteP{evaluating use cases} or how many years the organization had already models in production. This is commonly referred to as an organization's AI maturity\footnote{\url{https://info.algorithmia.com/hubfs/2019/Whitepapers/The-State-of-Enterprise-ML-2020/Algorithmia_2020_State_of_Enterprise_ML.pdf}, page 8.}. We also asked participants for goals within their organizations ML-model checklist~\cite{samek2017explainable, miller1968response}.

\myparagraph{Questionnaire--demographics} The last part of the survey was about the individual background of our participants. It covered basic demographic questions, but also a description of participants' roles within their teams to address practitioners with formal and instrumental knowledge~\cite{suresh2021beyond}, and to include a broader AI audience~\cite{arrieta2020explainable}. To test for a possible relevance of field related expertise, we also queried participants' self-reported knowledge in ML and asked whether they had taken any lecture in ML, security or AML~\cite{bieringer2021mental}.

\begin{figure*}[!ht]
 \centering
    \begin{minipage}{0.31\textwidth}
    \centering
      \includegraphics[  width=1.03\textwidth]{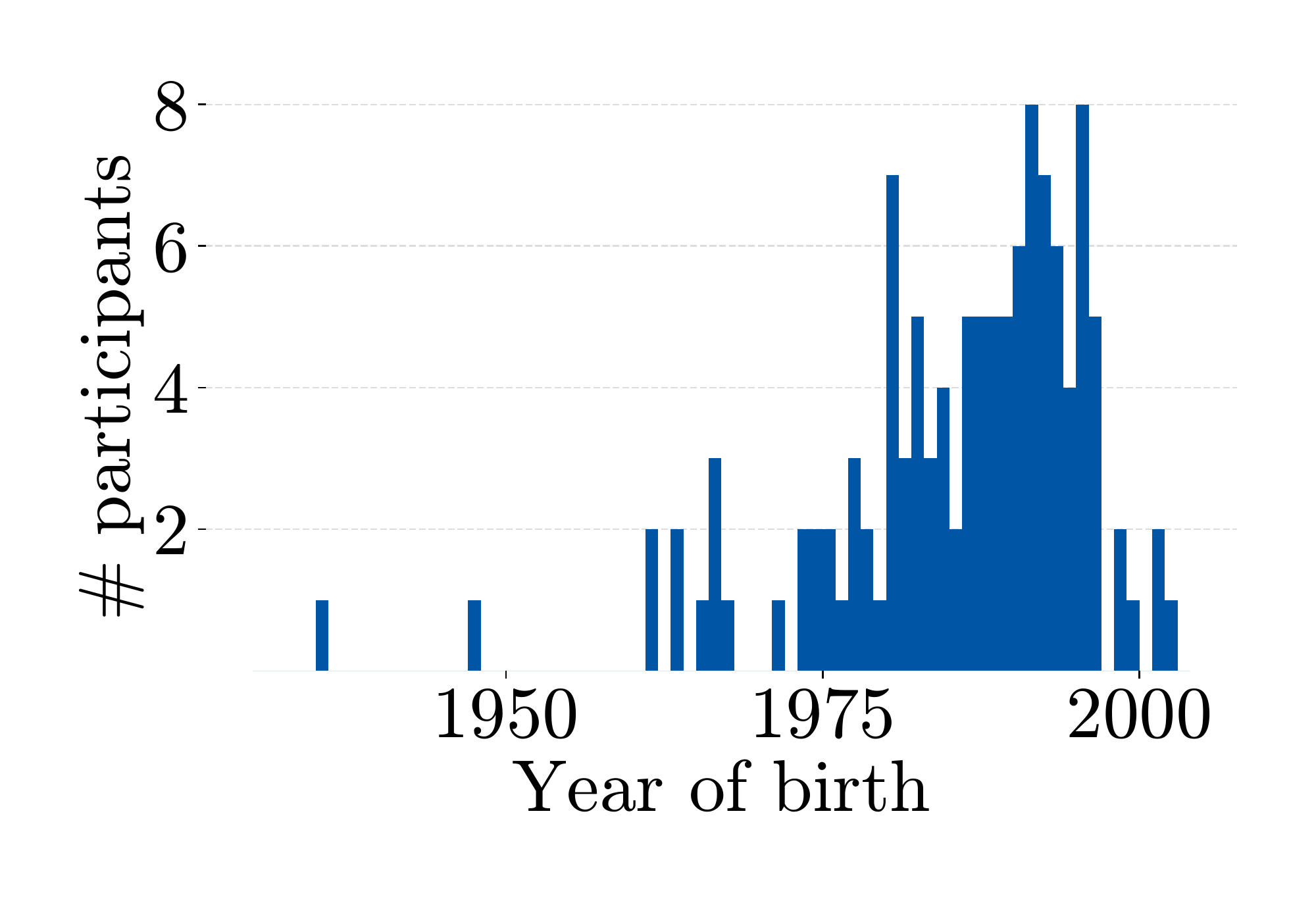}
      \caption{The age distribution (years of birth) of our participants.}\label{fig:ages}
    \end{minipage}%
    \hfill
    \begin{minipage}{0.31\textwidth}
      \includegraphics[trim={0 0 0 31},clip,  width=1.03\textwidth]{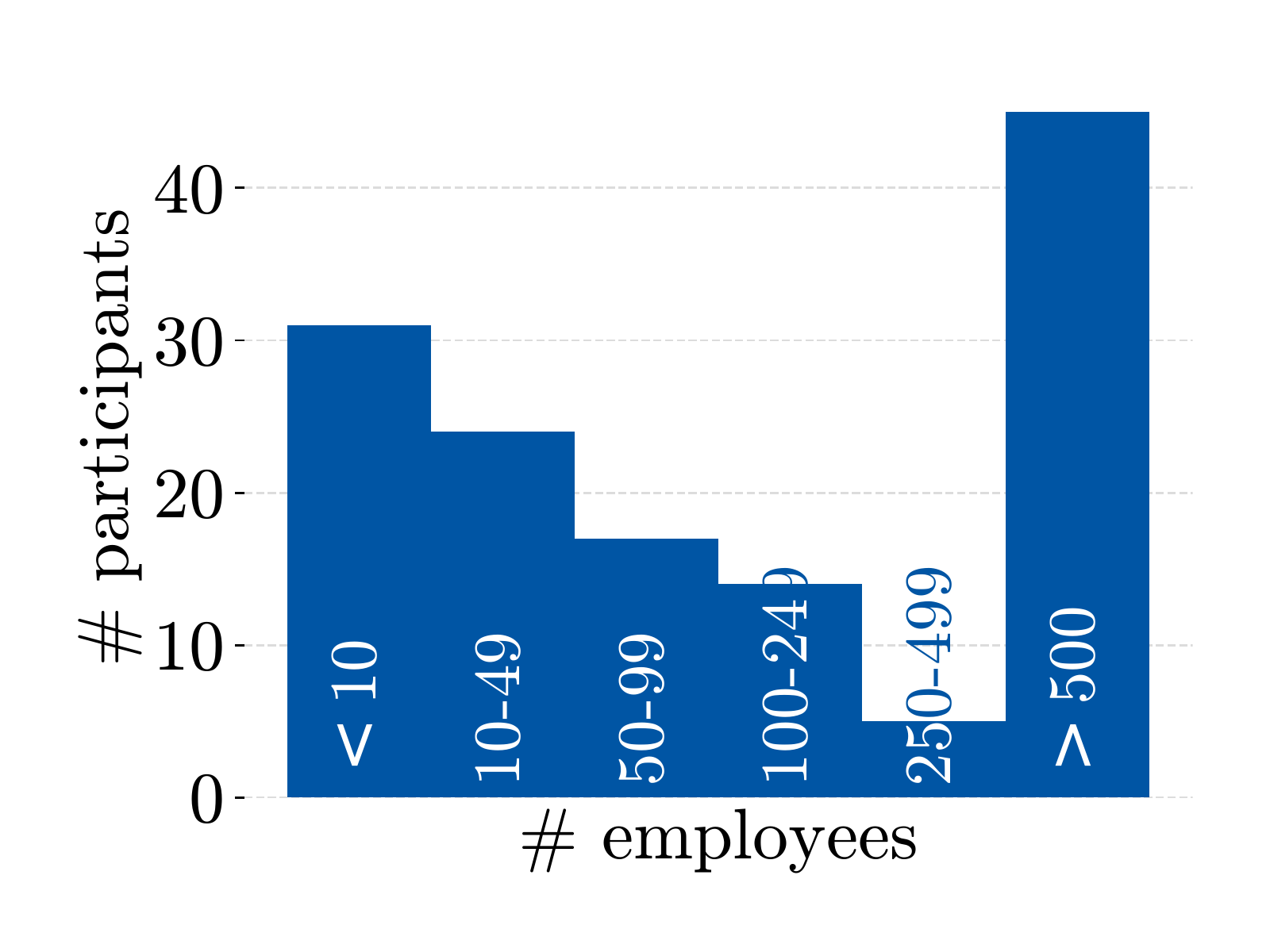}
      \caption{Grouping  our participants' organizations according to size.}\label{fig:sizes}
    \end{minipage}%
    \hfill
    \begin{minipage}{0.31\textwidth}
      \includegraphics[trim={0 0 0 31},clip, width=1.03\textwidth]{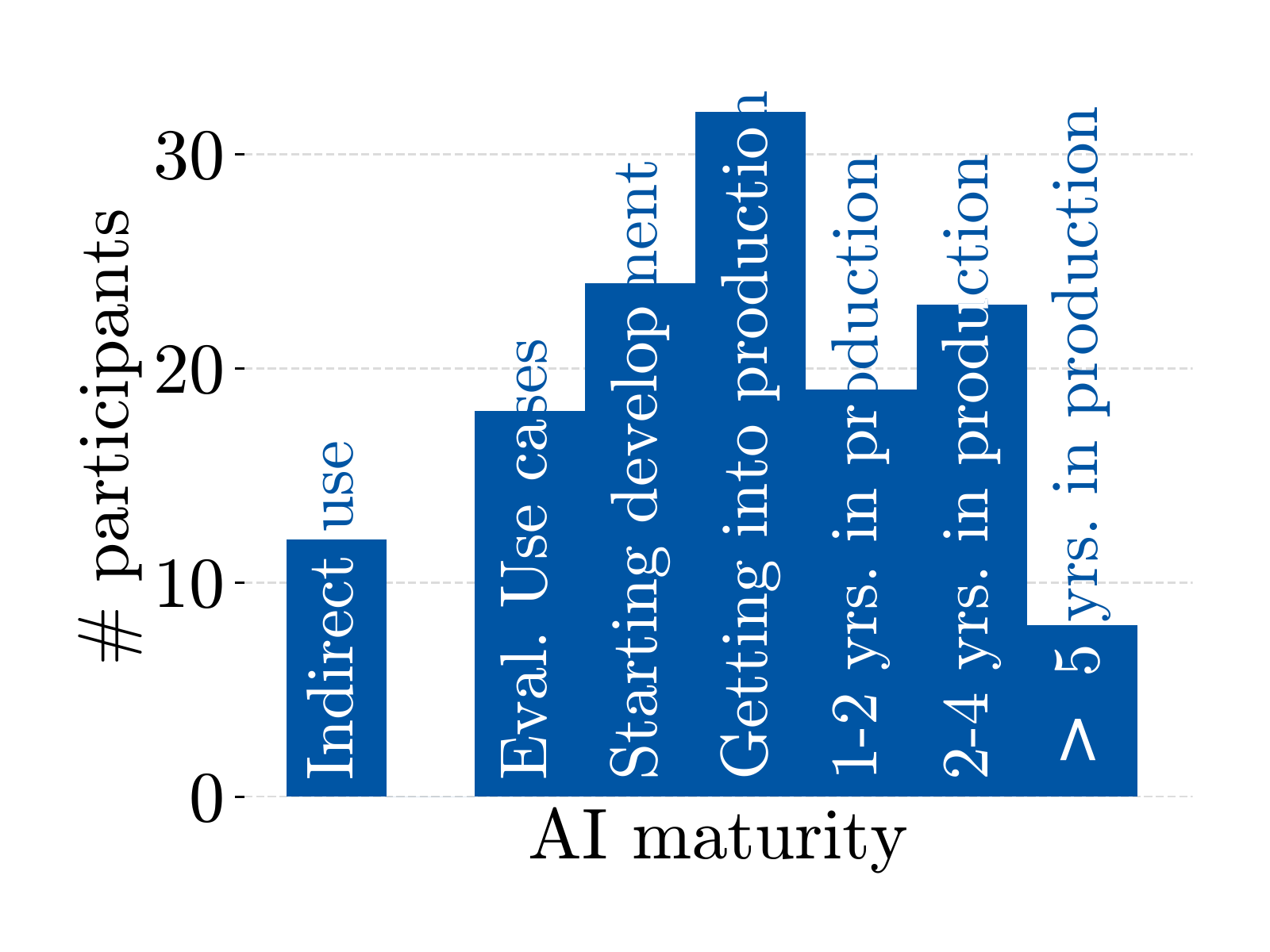}
      \caption{Self-reported AI maturity of participants' organizations.}\label{fig:maturity}
    \end{minipage}
    \hfill
\end{figure*}

\subsection{Pretests and recruiting} We implemented the questionnaire using Google Forms and ran a total of four rounds of pretests once there was the initial version of our questionnaire. The first three rounds with in total eight participants encompassed the full questionnaire. In the final round with three participants, we double-checked wording of some questions that were not sufficiently clear in the previous pretests. In this last round of feedback, no more necessary changes for the questionnaire emerged.
Once pretests had been completed and the final questionnaire implemented, we started recruiting participants in the direct network of the first two authors of this paper. In doing so, we aimed to enable any necessary final adjustments to the questionnaire itself and to the way we approached participants before the study was widely advertised on social media channels.

However, we found that direct messaging to both known and unknown possible participants came with higher conversion rates than general social media postings. Therefore, we joined several online communities for ML practitioners (e.g., R-Team for Data Analysis, Watson Developer Community, adversarial robustness toolbox, Data.Talks.Club) to approach potential participants via direct message on Slack. In doing so, we continuously monitored our sample with regard to representativeness to the overall target population. For example, the initial share of female participants in our study was below reported shares, and we therefore explicitly targeted female communities. Our initial power analysis required more than 97 participants. Taking into account that not all participants reply to all questions, and having reached 104 participants after two and a half months, the authors decided to recruit $>$125, yielding the eventual sample size of 139.

\section{Sample and data pre-processing}
We now discuss our sample encompassing the $139$ participants and the pre-processing of the free-text replies. 
We first review the individual background of our participants, and discuss gender, age, education and the professional role. We then focus on the organizational background, and discuss organization areas, organization location, size, and concrete AI usage.
Afterwards, we discuss the detailed procedure how we analysed and encoded the free-text replies and the agreement that we obtained across the two coders.

\subsection{Sample description}\label{sec:sampledescr}
A total of 139 participants filled our questionnaire, with additional 5 participants submitting empty forms (total 144). In addition, nine  participants whom we contacted reported to have had a look at the questionnaire but did not want to participate as they felt they did not have enough knowledge, had not been exposed to the topic or felt the topic was not relevant in their area. One additional participant denied to take part due to confidentiality reasons. 

\myparagraph{Individual background of participants}
Of our $139$ participants, more than two-thirds (71.2\%) were male, 14.4\% female, the remainder did not reply or did not want to disclose their gender. Albeit the sample is largely male, the percentage of female participants is comparable to reports studying the larger ML practitioner population~\cite{kaggle}. The distribution of participants’ year of birth was mostly between 1974 and 1996 (median 1986, see Fig.~\ref{fig:ages}), and is also similar~\cite{kaggle}. Also the distribution of academic degrees, with the largest group of master degrees (45.3\%) roughly mirrors this distribution~\cite{kaggle}. Beyond general education, only few participants self-reported none or little knowledge in ML (5.7\%). Many reported moderate (39.5\%) or high knowledge (40\%). More specifically, in a question asking for ML, AML and security knowledge, over three quarters of our participants reported ML knowledge (82\%), only a third reported to be knowledgeable in security (34.5\%) and even less in AML (28.7\%). Less than a fifth reported knowledge in all three areas (17.3\%). The most frequent combination was ML and security (30.2\%), then AML and ML (28\%), then AML and security (18\%). 
Finally, the most frequent role in team is ML engineer, which almost a fourth (23.7\%) indicated. The two second largest groups, both roughly one fifth, were ML researchers (20.9\%) and data scientists (20.1\%). Our sample also encompasses a few rather less technical roles, including nine domain experts (6.5\%), five auditors (3.5\%), and three product owners (2\%). Roughly a fifth of the  participants (18\%) preferred to specify their own role, and named for example roles like \quoteP{consumer}, \quoteP{technologist}, \quoteP{consultant}, or \quoteP{CEO}. 

\myparagraph{Organizational background of participants}
These roles were filled in organizations of overall diverse areas, with the largest two groups being healthcare (12.9\%) and IT security (10.8\%). Further areas included, but were not limited to marketing (6.5\%), computer vision (5.7\%), and finance and insurance (5\%). 
Albeit most of the participants’ organizations were located in US and Europe (69\%), our survey covered organizations from at least 26 countries (roughly nine out of ten participants (87\%) provided a country). 
More than a fifth of organizations participating in this study had 1-49 employees (22\%) while 45 (32.3\%) participants were working for organizations with more than 500 employees (see also Fig.~\ref{fig:sizes}), similar to existing industry surveys in ML~\cite{kaggle}.  
With regards to the AI maturity of their organizations, 24 (17.2\%) participants stated that their organization was starting to develop models whereas 32 (23\%) reported that their organization was getting developed models into production. Albeit 42 (36.6\%) participants reported that their organization had models in production for 1-4 years, relatively few subjects (5.7\%) reported that their organization had ML models in production for more than 5 years (see Figure~\ref{fig:maturity}). 
Concerning the concrete usage of ML techniques,
more than half of our participants stated they used supervised learning (56.8\%). Significantly less, about a fifth, used semi-supervised learning (19.4\%), less unsupervised learning (14.4\%) and few reinforcement learning (6.5\%). Even fewer reported to work with all four categories of learning (3\%).
Related to the previous question, two thirds of our participants used categorical labels (67.6), roughly half structured labels like bounding boxes (51\%) or real valued labels (49\%). Less worked with unlabelled data (41.7\%), and almost a fifth reported to work with all kinds of labels (18\%).

\subsection{Data pre-processing}
Our questionnaire encompassed several possibilities for participants to reply with free text, for example in the question about dreaded threats or threat relevance questions.
To analyse these replies, the first two authors of this paper applied four rounds of open coding. In each round, each coder assigned one or several codes to each participants statements, which were then discussed alongside with open or arising questions. We then performed Strauss and Corbin’s descriptive axial coding to group our data into categories and selective coding to relate these categories to our research questions~\cite{strauss1990basics}. Throughout the  coding  process,  we  used  analytic  memos  to  keep  track of thoughts about emerging themes. The final sets of codes are listed in App.~\ref{app:codes}.

\begin{table}[]
\centering
\caption{Inter-coder agreement of text replies. We compute Spearman correlation ($^*$) and Cohens Kappa ($^\dagger$) for the most feared threat (\question{1}) and the replies of high/low relevance of the four investigated AML attacks. Total codes refers to the maximum number of codes given from either coder. }\label{tab:interratorAgreement}
\begin{tabular}{llrrrrr}
\toprule
 
 &&& Agree- & Total  & \# dis-  & \\
 &&& ment & codes & agreeing  & \# replies \\
 \midrule
 \question{1} & \multicolumn{2}{r}{Most feared threat} & $.96^*$ & 232 & 37  & 136 \\  
 \specialrule{0.25pt}{0.5pt}{0.5pt}
 && high & $.79^\dagger$ & 86 & 6  & 69 \\ 
\multirow{-2}{*}{\question{8}} & \multirow{-2}{*}{Poisoning} & low &
$.65^\dagger$ & 61 & 8  & 51 \\ 
 \addlinespace
 && high & $.77^\dagger$ & 67 & 7 & 54 \\ 
\multirow{-2}{*}{\question{10}}&\multirow{-2}{*}{Evasion} & low & $.69^\dagger$ & 63 & 8  & 49 \\ 
\addlinespace
& & high & $.67^\dagger$ & 54 & 8  & 45 \\ 
\multirow{-2}{*}{\question{16}}&\multirow{-2}{*}{Membership} & low & $.53^\dagger$ & 47 & 11 & 45 \\  
\addlinespace
 && high & $.62^\dagger$ & 53 & 10  & 45 \\ 
\multirow{-2}{*}{\question{18}}&\multirow{-2}{*}{M. Stealing} & low & $.55^\dagger$ & 47 & 9   & 40 \\ \bottomrule
\end{tabular}
\end{table}

After coding, we computed annotator agreement. 
Given one document with many small text fragments, we opted for the Spearman correlation coefficient as a measure for annotator agreement~\cite{mcdonald2019reliability,jinyuan2016correlation} for the questions about most concerning threats. This correlation coefficient, while not encompassing random overlap, allowed us to take into account how often each code was used within the single document. For the relevance coding, we instead computed Cohen's kappa~\cite{cohen1960coefficient}, as we encoded high and low relevance for each of the five attacks separately, yielding several documents with varying code assignments. 
We report the detailed agreement measures and code numbers in Table~\ref{tab:interratorAgreement}. We drop one attack, \question{12}, as it is similar to poisoning and does not yield additional insights.
Given the semi-technical nature of our codebook, we consider our agreement substantial. In the following, we refer to the number of codes assigned in agreement by both coders.

\begin{table}[]
\centering
\caption{Summary of tested statistical hypothesis and findings. Primary variables are in the columns, and are tested against secondary variables in the rows. A \checkmark denotes that the relationship is statistically significant, if there are brackets, some attacks are statistically significant. X\textcolor{white}{,} means there relationship is not statistically significant and $^*$ denotes results which are not discussed in this paper.}\label{tab:resTable}
\begin{tabular}{lrrrr}
\toprule
 & Estimated & Threat  & Number of  &  Attack\\
 & exposure & exposure & mitigations  & concern \\
 \midrule
 Organizational Area &X$^*$ & X\textcolor{white}{,} &X$^*$ & (\checkmark) \\
 AI maturity &X$^*$ & X\textcolor{white}{,} & \checkmark\textcolor{white}{,} &X$^*$ \\
 Company size &X$^*$ & X\textcolor{white}{,} & X\textcolor{white}{,} &X$^*$ \\
 \addlinespace
 Techn./non techn. & X\textcolor{white}{,} & X\textcolor{white}{,} &X$^*$ & X\textcolor{white}{,} \\
 AML knowledge & X\textcolor{white}{,} & X\textcolor{white}{,} & \checkmark\textcolor{white}{,} & \checkmark\textcolor{white}{,} \\
 ML knowledge & X\textcolor{white}{,} & X\textcolor{white}{,} &X$^*$ & (\checkmark) \\
 \addlinespace
 Estimated Exposure & &X$^*$ & \checkmark\textcolor{white}{,} & (\checkmark) \\
 Threat Exposure &X$^*$ & & \checkmark\textcolor{white}{,} & X\textcolor{white}{,}\\
\bottomrule
\end{tabular}
\end{table}

\section{Results}
We now report the results of our study by
analyzing our participants' responses and running statistical tests. The detailed statistical results including all test and sample statistics can be found in App.~\ref{app:stattests}. 
We analyze our sample on three layers that emerge in the context of our research questions. The first one are insights related to the occurrence of attacks in practice, the second influential factors and risks at the organizational respectively third at the individual level. 

\myparagraph{AML attacks in practice} 
We find evidence for AML circumventions (Sect.~\ref{sec:AMLQTwo}). Although some participants are concerned about AML and
name corresponding threats explicitly (Sect.~\ref{sec:amlQOne}), we also find that \nonMLSecurity (e.g. access control, botnets, resource theft, etc), and privacy are encountered threats and concerns. Finally, participants name generic ML (i.e., data drift) and organizational (i.e., security awareness) challenges.

\myparagraph{AML within organizations}
In Sect.~\ref{sec:orgaStatTests}, we find no statistical significant predictors for threat exposure. At the other hand, the organization area does influence risk perception and implemented mitigation depends on exposure and concern. We also find that the reasons for threat concern are highly complex (Sect.~\ref{sec:prgaRelevance}): Our participants consider both a wider range of impacts of an attack (financial or business, up to ethics) when deciding if an attack is relevant. We find that furthermore, when an attack is perceived as irrelevant, often it is described as infeasible given deployment or use-case, or usage of the model. However, in some cases, attacks are rather seen as benign failure cases.

\myparagraph{AML for practitioners}
In Sect.~\ref{sec:practitioners}, we find that prior knowledge (largely in AML, but also in ML) leads to a higher concern about individual threats. Furthermore, gender sometimes influences threat perception, but not consistently.

We give a summary of all statistical results in Table~\ref{tab:resTable}. 

\subsection{AML attacks in practice}
We start with the discussion of witnessed attacks and then discuss  the dreaded attacks by our participants.

\subsubsection{Encountered AML threats}\label{sec:AMLQTwo}
We first consider the estimated likelihood to become victim of an attack (\question{3}). Less than a fifth (17.2\%) of our participants estimate the likelihood of an AML attack within the next 12 months as high or very high. Instead, roughly half (49.6\%) estimate the likelihood as low or very low---indicating that exposure might not be very high. 
Hence, we had asked participants whether they had encountered a circumvention of their AI based workflows or systems (\question{2}). This was confirmed by less than a fifth (17.3\%) of our participants. More concretely, seven participants (5\%) witnessed one circumvention, six (4.3\%) two, one (0.5\%) three, two participants (1.4\%) four circumventions and eight (5.7\%) more than four.
To obtain more in depth knowledge, we asked participants to briefly describe the circumvention in a free text field, which we now discuss.

\myparagraph{AML in the wild} Of all  replies, five (3.6\%) were AML threats. Three (2.1\%) described an evasion attack. The first reply was in relation to HR (\quoteP{users spam to optimize their strategy for job search}),
the second two related to autonomous vehicles (\quoteP{autonomous vehicle image recognition errors leading to dangerous path planning}). In the case of the latter two, participants doubted \quoteP{an 'intentional' circumvention.}
Furthermore, there were two (1.4\%) cases of poisoning. Whereas one remains vague, writing about \quoteP{ML systems being retrained to  provide false outputs}, the second one was very detailed, reporting that \quoteP{partner employees tasked with labeling training data feel threatened by automation, and either stall or sabotage the labeling effort, harming the models.}

\myparagraph{Unclear replies} Further nine replies (6.5\%) contained no text, or replies like \quoteP{no details} or \quoteP{brute force attacks}, that do not allow to deduce the exact nature of the circumvention. An additional six replies (4.3\%) were  data breaches. Whereas some referred on a high level to \quoteP{data privacy} or \quoteP{incorrect data access,} others were slightly more detailed: \quoteP{acquiring the data for training AI systems}). In these cases, we assume, but cannot be sure, that they are not ML related.

\myparagraph{Circumventions not directly related to ML}
In total four (2.8\%) descriptions were not ML related security threats, including resource theft (2, i.e., \quoteP{we got hit by crypto-miners pretty hard [...]}), man-in-the-middle attacks (1, \quoteP{a man in the middle attack between two workflows [...]}) and botnets (1, \quoteP{botnet communication}).

\myparagraph{Attacks mentioned in relevance reasoning}
We later inquired about the relevance of specific AML attacks. In these replies, some participants reasoned that they had witnessed the threat already. 
One participant wrote for example, in the context of poisoning, \quoteP{however, something kind of like a poisoning attack happened, but was because of an over-prevalent family of malware that warped the model into performing worse than the last one. This did impact the deployment, but was because of a poorly configured filter not an attack.}
Another participant reported to \quoteP{[...] [have] evidenced during a penetration test scenario} poisoning, and evasion. 
Another participant reports in the context of membership inference: \quoteP{we have seen users try to figure out what content will trigger our different abuse and spam identification models by trying different comment inputs and sharing these thoughts with others to help them bypass the potential identification.} 

\myparagraph{Conclusion}
There were occurrences of ML attacks in practice, namely poisoning and evasion. However, it was not always clear whether circumventions are \security or \safety issues, in other words benign or attacker based failures.
Furthermore, almost a third of our participants' replies remained vague, not allowing to understand the exact nature of an attack. Almost another third of replies were data breaches, privacy leaks, or other \nonMLSecurity issues.

\subsubsection{Concerns about AML}\label{sec:amlQOne}
We further aimed to understand what AML challenges practitioners face (\question{1}). To avoid priming, we had asked this before mentioning any specific AML attack. Of all participants, almost all (93.5\%) provided a reply, and more than a fifth (22.9\%) provided more than one concern. In the following text, as more than one code could be assigned to a reply and the total number of assigned codes is not equal to the number of participants, we report no percentages. 

We tagged 21 times security challenges that were directly related to the AML, for example \quoteP{data poisoning} or \quoteP{understanding the threats and associated risks of AI (and especially ML) - specific attack.} Several concrete AML attacks we later queried about, including poisoning (7), evasion (3), and model stealing (1) were named by our participants. 
However, most replies did not (only) contain AML threats. A few challenges, 10, were related to ML, for example \quoteP{explainable ML/NN} or \quoteP{concept drift}.
There were also 16 challenges related to privacy. These encompassed \quoteP{data protection, legal data collection, GDPR, information security}, in other words both general privacy (10) concerns as also the challenge to be compliant with legislation (6).  
Several (20) challenges concerned security in organizations. Corresponding replies are for example \quoteP{convincing stakeholders of the risks}, \quoteP{protecting intellectual property} or \quoteP{achieving security guarantees while reducing false-positives}. They outline that challenges in AI can also encompass communication of risks (8), protecting intellectual property (7) or trade-offs that arise when both several factors are balances against each other (4).
Furthermore, there were 35 challenges related to \nonMLSecurity, including \quoteP{user access control}, or \quoteP{open source supply chain (ie - NPM / Log4J vulnerabilities)}. One participant reasoned: \quoteP{hard to say but the traditional cybersecurity attacks are generally applicable in AI and those still seem to be most prevalent. [...] The adversarial scenarios as presented by evasion or poisoning are not as prevalent}, thus explaining why these replies are not about AML although we explicitly asked about it.
The largest used group of assigned codes (52) was related to data. While some of these replies were vague (11, \quoteP{data leak}; 17, \quoteP{data security}), some were related to sensitive data (17, \quoteP{PHI/HIPPA}) or challenges when sharing data (8, \quoteP{the biggest difficulty is safely sharing data with others}). In theory, almost all AML threats can be seen as attacks through data (through training for poisoning, 
through test data for evasion, membership inference and model stealing). However, threats caused by data could also include \nonMLSecurity, data quality, privacy, etc. We thus leave more detailed research on this question for future work.

\myparagraph{Conclusion} There were few, but some concrete mentions of AML. Although we had explicitly asked for AML, participants also raised \nonMLSecurity and privacy concerns, reasoning that these were more pressing than AML. Concerns also encompassed organizational challenges related to ML itself or risk communication or assessment. Finally, participants often reasoned vaguely about data security, leaving open whether they refer to data quality, privacy, or AML issues.

\subsubsection{Conclusion}
We find that AML threats did occur in practice, and that some participants were explicitly concerned about AML. At the same time, it remains sometimes (for example in evasion) unclear whether an incident is a \security or \safety issue. Furthermore,
concerns encompass \nonMLSecurity, privacy, organizational challenges and ML problems such as dataset shift, for example. 

\subsection{AML within organizations}\label{sec:Orga}
In this subsection, we examine AML in practice from an organizational perspective. To this end, we first relate different questions via statistical tests, and then 
analyze the arguments about individual attack relevance from our participants.

\subsubsection{Organizations approaches to ML security}\label{sec:orgaStatTests}
We first analyze whether the organization area influences threat perception, and then attempt (but fail) to find factors from our questionnaire that predict threat exposure. Finally, we  investigate which factors influence the implementation of mitigations.

\myparagraph{Organization area and threat perception} We assumed that the area an organization operates in affects threat perception. For example healthcare is based on sensitive data, thus health-care workers may be more concerned about related threats, in this case membership inference.
Our sample contains two large industry groups (\question{25}): IT security (15 participants) and healthcare (18 participants). We tested for both groups whether there were threats perceived as more relevant compared to the rest of the sample.
We divided the sample into one subgroup fulfilling the criteria and the rest of the sample, and used a Mann-Whitney-U test to determine if concern deviated in a statistical significant manner. For healthcare, we investigated membership inference but found no statistical significance ($p=0.7$). In the case of security companies, the relevance of backdoor attacks was statistically significant ($p=0.002$), as well as
evasion ($p=0.02$) and membership inference ($p=0.026$). Other threats did not exhibit statistical significance. 

\myparagraph{Predicting threat exposure} When it comes to threat exposure, we assumed that both the organization area plays a role as well as the amount of exposed AI technology of the organization and its visibility.
In our questionnaire, these were the organization area as defined in the previous paragraph, AI maturity (\question{24}) and organization size (\question{18}). The organization area was tested with a Mann-Whitney-U test, the latter two with an ordinal regression model. We did not find any statistically significant relations.  

\myparagraph{Predicting threat concern}
We expected that the individual threat concern may depend on both threat exposure (\question{2}) and estimated exposure (\question{3}), and thus tested both with an ordinal regression model. In terms of threat exposure, we could find no statistical significant relationship to any attack. In case of estimated exposure, we do find that it statistically significantly predicts both concern in case of poisoning ($p=0.002$) and evasion ($p=0.003$).

\myparagraph{Predicting the amount of implemented mitigations} We asked our participants about the number of implemented mitigations (\question{4}, for example approaches like \quoteP{documentation}, \quoteP{fail safe plans}, or \quoteP{incident response}).
We assumed that the implementation of these mitigations depends on factors such as previous exposure to threats, estimated risk to become  victim of an attack, organization size  (how much personnel can be dedicated to securing models), or how long models are in production. More concretely, we tested if the number of implemented mitigations (0-7) was influenced by factors like
exposure to threats (\question{2}),
estimated risk to become a target of an AI circumvention (\question{3}), organization size (\question{18}), and
AI maturity (\question{25}). We used an ordinal regression to model these relationships and found that exposure ($p=0.012$), estimated risk ($p=0.013$) and AI maturity ($p=0.004$) were statistically significant predictors. 

\myparagraph{Conclusion} We found that while the organization area affects the perception of some threats significantly, there were no statistically significant variables for threat exposure. The amount of implemented mitigations was however statistically related to threat exposure, estimated risk, and AI maturity.

\subsubsection{Concern about AML threats}\label{sec:prgaRelevance}
\begin{figure*}
    \centering
    \includegraphics[width=0.99\textwidth]{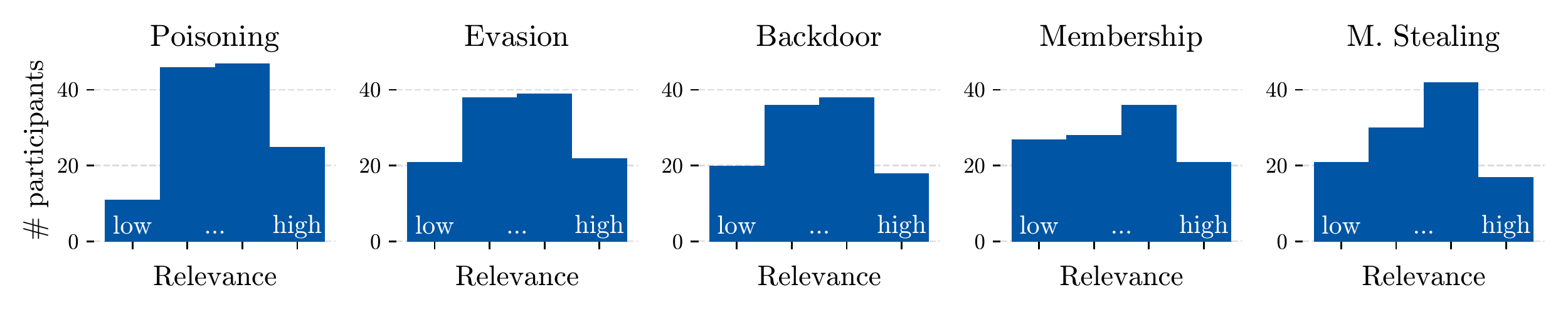}
    \caption{Reported relevance for the all five attacks we tested. The  Likert-scale provided had four items ranging from \quoteP{irrelevant} to \quoteP{very relevant.}}
    \label{fig:relevance}
\end{figure*}

In this subsection, we analyze the arguments provided by our participants when reasoning that a threat is relevant or irrelevant.
Previous work studied factors on threat concern such as the ease to attack and defend, or possible benefit of carrying out the attack~\cite{mirsky2021threat}. We instead asked our participants without priming to give a short reason for the relevance or irrelevance of an AML threat given a two sentence description (but not the name) of the attack. More concretely,
we asked our participants how relevant they thought poisoning (\question{8}), evasion (\question{10}),  membership inference (\question{16}) and model stealing (\question{18}) was. In this version of the paper, we do not discuss backdoors (\question{12}), as they are similar to poisoning. We also asked about one additional sanity-check threat (\quoteP{altering training data to delete an untrained model. In other words, the training data contains a pattern that will delete the model after training.}, \question{14}). Although some participants reported high concern, the threat was rated statistically significantly less relevant compared to all other threats\footnote{Mann-Whitney-U test with $[1.4e^{-10}< p < 1.2e^{-16}]$.}. We thus omit the sanity threat in the following discussion, where we first discuss the high relevance of each poisoning, evasion, 
membership inference, and model stealing. The same order is used for the discussion of irrelevance replies. A summary of our results is depicted in Table~\ref{tab:relevanceCoding}, and we plot the numerical relevance ratings in Figure~\ref{fig:relevance}.

\myparagraph{Poisoning--high relevance}
The most frequently coded reply reasoning for relevance, occurring 10 times, was the relevance within the applications setting of the participant (\quoteP{we use AI for security purposes, tampered training data is one of the best ways for attackers to evade the system}). Following up codes were associated with relevance without argument (9, \quoteP{yes}), and two codes associated with model performance (9 and 9). Participants also reasoned that an attacker was credible (5, \quoteP{sharing data across multiple users makes this a threat that needs to be considered}), or that they understood the attack (7). Finally, some participants reported  exposure to the attack (3), which is rarely the case for other attacks.

Furthermore, we found that 4 times, participants found the threat relevant as it would cause wrong decision making (\quoteP{models inform our decisions. Wrong models imply wrong decisions.}). They furthermore reasoned that poisoning caused financial loss (3, \quoteP{altering training data could result [...] in catastrophic increased spending}) for their organization or harmed fairness by potentially introducing bias (3).

\myparagraph{Evasion--high relevance}
The most frequent reply for high relevance of evasion was impact on model performance (11 times). At the same time, 6 participants reasoned that although evasion is relevant, it is not a security issue (6 times, \quoteP{it may be a case of overfitting}). Further reasons included that evasion was easy to carry out (4), hard to defend (4), a threat relevant in the given application (3, \quoteP{attackers targeting our systems in this way may break them}), or assumed to be relevant without providing an argument (4, \quoteP{it is}).

As in poisoning, participants also reasoned that evasion affects decision making in their companies (4), or negatively affects fairness, bias, or ethics (3, \quoteP{brings in bias}).

\myparagraph{Membership inference--high relevance} Most participants argued that they were concerned about the resulting data breach (21, \quoteP{the possibility of de-anonymizing data would be a concern that can’t be understated}). Some participants understood the underlying mechanism (4, \quoteP{it allows someone to reverse engineer the inputs and potentially identify where the data came from as well as who or what is/isn’t included}), other reasoned that the threat was relevant in their specific use-case (3, \quoteP{especially our model could be queried to generate training data}) or did not give additional arguments (3). 

Our participants also reasoned that membership inference causes business information leakage (3, \quoteP{could be relevant because it would allow our clients to get information about the competition they would normally not have.}) or noncompliance with existing regulations (3, \quoteP{GDPR requires that I don’t accidentally leak data that was supposed to remain private}).

\myparagraph{Model stealing--high relevance} 
Most participants stated that model stealing results in a loss of their intellectual property (8, \quoteP{stealing IP}). 
Further, participants reasoned that the attack was easy to do (4), was relevant in their application setting (3, \quoteP{it might lead to our models being reverse-engineered by clients.}) or the attacker had a motivation to carry out the attacks (3, \quoteP{when scraping enough data one could probably “copy” our models}). Practitioners also reasoned based on their understanding of the attack (4, \quoteP{technically its no brainer - it's very much possible}).

Compared to other attacks, much more participants remark on the impact of model stealing. Several participants mention general business consequences (5, \quoteP{threat to the business}), whereas others address profit for a competitor (5, \quoteP{would allow competitors to achieve our better results with minimum efforts.}), financial loss (4, \quoteP{it costs a lot of money to train giant networks, hence the problem is very relevant in terms of investment}), and business information leakage (3, \quoteP{could give unfair insights in our decision making}).

\myparagraph{Poisoning--low relevance}
The most frequent code (14) for irrelevance of poisoning attacks was that the data was not accessible to 3rd parties or the outside of the organization (\quoteP{no one can access the training samples}). Additional frequent codes were that the threat is not relevant under the use case (9 times, \quoteP{our training data comes [...] from clinical studies we conduct ourselves [...] so chances that someone interfere with the data gathering process are very low}) or doubting the attacker (8, \quoteP{we do not think any actor would be sufficiently motivated to attempt it}).
While some participants stated that their human in loop (5, \quoteP{the training data is curated by us}) or another defense they implemented (5, \quoteP{very few publicly available data used for training}) prevented the attack. Finally, some (3) also reason that the attack is hard to carry out.

\myparagraph{Evasion--low relevance} Most participants (11) arguing against the relevance of evasion denied that an attacker could access the required test data. Almost as many reasoned based on their specific use cases (10, \quoteP{the podcast audio is stored with a number of distributors [...]. The corruption would have to occur amongst multiple distributors [...].}). Many participants also doubted the attacker's motivation (7, \quoteP{[...] there would not be enough benefit to the actor}). Further reasons included that the attack was hard to do (3), or that a defense was implemented (4, \quoteP{[...] the attack surface to alter data is minimized by multifactor access, role based access controls, time based tokens, logging, monitoring, and encryption.}). Finally, we tagged some replies (5) as confused threat models because participants referenced training data (\quoteP{training data is usually high quality}). 

\myparagraph{Membership inference--low relevance} To reason for the irrelevance of membership inference, participants often referred to their specific use case (10, \quoteP{we work on new data in news and the probability of that happening since our models are trained in old data is very unlikely}) or directly stated they were dealing with non-sensitive data (9, \quoteP{the training data is publicly available anyway}). In addition, participants sometimes did not provide an additional argument (4), doubted the attacker (3, \quoteP{for our use cases, I can’t (yet) see how anyone would stand to gain from this}) or reasoned their model was not accessible at test time (3, \quoteP{the model cannot be queried directly by the users}). 

\myparagraph{Model stealing--low relevance} 
Most participants (13) that deemed model stealing irrelevant reasoned based on their use-case (\quoteP{the use of the model requires domain knowledge so it’s unlikely that someone outside the organization would be able to make a correct interpretation of it’s functionality}). Otherwise, participants remarked that their models were not accessible (7, \quoteP{we don’t offer API’s to our models.}), or were replaced often and copying them yielded no benefit (3, \quoteP{model is continuously updated, and previous models don’t have much value}). 
Participants also reasoned that the attack was hard to carry out (4), or generally irrelevant (3, \quoteP{this is a business model issue, not a technical issue}). Participants also doubted that an attacker might benefit (3, \quoteP{the value of copying [our models] would be quite small for someone else}), or reasoned that the attack does not apply in their use case or way to deploy ML (3, \quoteP{the model is likely to be deployed on edge devices so it will be anyway known to the potential attacker.}).


\begin{table}[]
\caption{Participants' argumentation for the relevance of attacks. For each attack we present the five most frequent 
(ties broken randomly) arguments (and their frequency).}\label{tab:relevanceCoding}
\begin{tabular}{@{}lrl@{}}
\toprule
  &   \textbf{RELEVANCE} & \textbf{IRRELEVANCE}  \\
  \cmidrule(r){2-2} \cmidrule(r){3-3}
  {\multirow{5}{*}{\rotvertical{Poisoning}}} 
  &  \fontfamily{cmss}{Relevant in application (10)}  & \fontfamily{cmss}{(14) No data access}\\
  & \fontfamily{cmss}{Impact on safety (9)} & \fontfamily{cmss}{(9) Not relevant in application}  \\
   & \fontfamily{cmss}{Impact on performance (9)} & \fontfamily{cmss}{(8) Doubting attacker}  \\
  & \fontfamily{cmss}{Relevance without argument (9)} & \fontfamily{cmss}{(5) Human in the loop defense}  \\
  & \fontfamily{cmss}{Hard to defend (7)} & \fontfamily{cmss}{(5) Some defense implemented}  \\
\addlinespace \cmidrule(r){2-2} \cmidrule(r){3-3}
  {\multirow{5}{*}{\rotvertical{Evasion}}}& \fontfamily{cmss}{Impact on model performance (11)}&\fontfamily{cmss}{(11) No data access}  \\
  & \fontfamily{cmss}{Impact on safety (6)} & \fontfamily{cmss}{(10) Not relevant in use case}  \\
   & \fontfamily{cmss}{Impact on decision making (4)} & \fontfamily{cmss}{(7) Doubting attacker}  \\
  &\fontfamily{cmss}{ Easy to do (4)} & \fontfamily{cmss}{(4) Some defense implemented} \\
  & \fontfamily{cmss}{Hard to defend (4)} & \fontfamily{cmss}{(3) Hard to do}  \\
\addlinespace \cmidrule(r){2-2} \cmidrule(r){3-3}
  {\multirow{5}{*}{\rotvertical{Membership}}} & \fontfamily{cmss}{Data breach (21)} & \fontfamily{cmss}{(10) Not relevant in use case}  \\
  & \fontfamily{cmss}{Relevance without argument (3)} & \fontfamily{cmss}{(9) No sensitive data}  \\
   & \fontfamily{cmss}{Business information leakage (3)} & \fontfamily{cmss}{(4) Irrelevance without argument}  \\
  & \fontfamily{cmss}{Regulatory compliance (3)} & \fontfamily{cmss}{(3) No query access}   \\
  & \fontfamily{cmss}{Relevant in application (3)} & \fontfamily{cmss}{(3) Doubting attacker}  \\
\addlinespace \cmidrule(r){2-2} \cmidrule(r){3-3}
 {\multirow{5}{*}{\rotvertical{M. Stealing}}} & \fontfamily{cmss}{Impact on intellectual property (8)} & \fontfamily{cmss}{(13) Not relevant in use case }  \\
  & \fontfamily{cmss}{General business impact (5)} & \fontfamily{cmss}{(7) No query access}  \\
   & \fontfamily{cmss}{Profit for competitor (5)} & \fontfamily{cmss}{(4) Hard to do}  \\
  & \fontfamily{cmss}{Financial loss (4)} &\fontfamily{cmss}{(3) Doubting attacker}  \\
  & \fontfamily{cmss}{Attacker credible (3)} & \fontfamily{cmss}{(3) Model shortlived} \\
\bottomrule
\end{tabular}
\end{table}

\myparagraph{Conclusion}
Our analysis shed light on the complexity of AML in practice. 
While poisoning was the only attack to be witnessed in practice by several participants, other threats were deemed relevant due to their (potential) impact. Such impact was very diverse, and ranged from
decreased model performance, 
wrong decision making, to business implications like leak of information, financial loss and leakage of intellectual property. 
When an attack was deemed irrelevant,  the attacker often would not have access to the required data. In this sense, both application and deployment are orthogonal factors influencing vulnerability: One use-case may be security critical only when deployed in a certain way, otherwise not. Finally, sometimes the difference between \security and \safety was not well distinguished.

\subsubsection{Conclusion}
Practitioners from IT security companies were significantly more concerned about AI threats. At the same time, in our sample, exposure to threats was not a factor of organization size or AI maturity. The amount of implemented mitigations did, on the other hand, depend on previous threat exposure, but also on expected risk, and AI maturity. 
Our participants, when reasoning about attack relevance, encompassed not only feasibility or the ease to mitigate an attack, but considered direct impacts like financial loss, information leakage, or business harm. Furthermore, decision making based on ML or practical encounters fuel concern. When threats were deemed irrelevant, this was usually based on specifics of the use-case or deployment and/or inaccessibility or the required resources or data for the attacker. Finally, in some cases, attacks were perceived as \safety  issues, or in other words benign failure cases.

\subsection{AML for practitioners}\label{sec:practitioners}
In this section, we investigate individual factors that may influence threat exposure, estimated threat exposure or threat perception. Such individual factors are the role in team, prior knowledge in AML, ML, prior education in general, or gender.

\myparagraph{Role in team} Assuming that an ML engineer is closer to model deployment than a product manager, we may assume that technical roles in teams are more exposed to threats. We thus investigated the reported role within a team (\question{30}) in relation to threat exposure (\question{2}), expected exposure, and  perceived attack relevance. To this end, we split our sample into technical (48 participants, for example \quoteP{ML Engineer}, \quoteP{ML Scientist}, \quoteP{data architect}, etc) and non-technical (91, for example \quoteP{product owner}, \quoteP{auditor},\quoteP{domain expert}, etc).
 We found, using a Mann-Whitney-U test, that threat exposure was not different for technical and non-technical roles ($p=0.15$), and there was no difference in expected exposure ($p=0.8$).
Furthermore, we tested using a Mann Whitney-U test whether the concern of these two groups differed statistically. For no threat, this was the case ($0.4<p<0.8$).

\myparagraph{Prior knowledge in AML}
Another possible individual factor is knowledge of AML. Knowledge means understanding, thus potentially raising threat concern and also motivating countermeasures or mitigations. In \question{32}, we asked our participants to self-report their knowledge in AML. We split our sample into two groups (knowledgeable 40, not knowledgeable 99) and tested for statistical significance in concern about threats, exposure to threats (\question{2}) and the number of mititgations implemented (\question{32}) using the Mann-Whitney-U test. 
There was no statistically significant difference between these two groups for threat exposure ($p=0.15$) or general threat concern ($p=0.5$).
However, participants who reported knowledge of AML were significantly more concerned about AML threats (except sanity, $0.006<p<0.018$).
We found finally that self-reported prior knowledge lead to a statistically significantly increased number of implemented mitigations ($p=0.0024$). 

\myparagraph{ML knowledge and education}
Given the high relevance of AML knowledge, we also tested the influence of general ML knowledge and education (e.g., Highschool, Bachelor, PhD) on both exposure (\question{2}), expected exposure (\question{3}), and threat concern. While we expected ML knowledge to have an influence on exposure and threat concern (most AML attacks are based on ML mechanisms), we did not expect general education to influence concern. We tested our hypotheses with an ordinal regression model, and found no statistical significance for exposure. For the individual threats, we found that only ML knowledge significantly influenced concern for evasion ($p=0.025$) and membership inference ($p=0.025$), but for no other threat ($p>0.5$). 

\myparagraph{Gender} We further investigated whether gender influences expected threat exposure or threat concern. We expected that women were similarly or more concerned, in line with previous findings~\cite{hossain2019security}. To test this hypothesis, we divided our sample into female (20 participants) and male (99 participants) and computed a Man-Whitney-U test on the replies for estimated exposure (\question{3}) and threat concern. Gender did not statistically influence the overall estimated risk ($p=0.43$).  
For most threats, there was no significant difference ($p>0.15$) either. However, for 
model stealing ($p=0.021$), we found that women are less concerned, contrary to our expectations. 
A possible explanation could be that women work with different applications and deployment settings, but our data did not allow to investigate this profoundly.

\myparagraph{Conclusion} In this subsection, we investigated factors specific to individual practitioners and their relation to threat exposure or perception. We find that prior knowledge (largely in AML, but also in ML) lead to a higher concern about individual threats. We furthermore find that gender influences threat perception, too, but not consistently.

\section{Limitations}
In this section, we discuss limitations in our study that affect the generalizability of our results. We first describe limitations within the sample, then within the questionnaire, and finally within the statistical approach.

\myparagraph{Sample limitations} 
Our sample is limited to English speaking practitioners, and biased towards the global north. Furthermore, we had initially planned to compare opinions from industry and academia, but got feedback early on that our questionnaire was too industry specific for academics. More specifically, some participating academics reported back to us to have filled the survey from the perspective of recent industry experience. In contrast to our expectations, the industry area does not allow to deduce which participants are from academia, making it hard to understand the influence academics could have had on the results. Due to the usage of different links to monitor recruiting strategies, however, we know that less than a quarter (35) of participants are pure academics without industry experience. Still, we might underestimate the occurrence of threats in the wild (Sect.~\ref{sec:AMLQTwo}). 

\myparagraph{Questionnaire limitations}
We did only consider self reported knowledge, and did not asses our participants' knowledge.
Independently, we failed to observe statistically significant results for type of learning (supervised, unsupervised), kind of labels (none, categorical, real), and type of data (vision, video, etc.). A possible cause for this, apart from the conclusion of indeed no  relation, is that participants for example work in several projects. This diversity was not foreseen by us when designing the questionnaire. 

\myparagraph{Methodological limitations}
We performed a sample size estimation upfront before designing the questionnaire (see Sect.~\ref{sec:methodology}). However, the Mann-Whitney-U test's sample size is dependent on factors such as mean and variances of both samples, factors that are impossible to approximate upfront. While in most cases, differences indicate that our sample size is enough, we cannot exclude that in some cases, the test returns a too conservative result (e.g., masking significance when there is indeed an effect in the data). These cases were when testing threat concerns for the role in team (Sect.~\ref{sec:practitioners}), and when testing whether healthcare workers are more concerned about threats like membership inference (Sect.~\ref{sec:orgaStatTests}). These effects should be re-evaluated in future work with a different design or more participants.

\section{Implications and future work}
Despite these limitations, our research yields practical implications for a better understanding of attack relevance, more granular risk assessments and an improved communication of AML risks. We now concretise these implications and conclude the paper by discussing open research questions that could be addressed by future work.

\myparagraph{Better understanding of general attack relevance} For each of the attacks we tested for, participants’ argumentation for relevance involved reasoning about the validity of
a threat in a certain application, deployment, or use case (Table~\ref{tab:relevanceCoding}).
More specifically, not all attacks apply in all applications or forms of deployment.
This finding is highly relevant for risk management in the real-world and implies that threat modelling should always consider the specific context of an AI system. To this regard, further research is needed to investigate which attacks should be considered in which application scenarios, and which deployment is prevalent in which application. Somewhat orthogonal, a quantification of the impact incurred by an attack (in terms of financial loss, for example) based on attack type and application would benefit a deeper understanding of risks related to ML in the wild.

\myparagraph{More granular risk assessments of AI systems} Our findings in Section~\ref{sec:Orga} help to understand which factors should be taken into account when threat modelling an AI application. This is relevant for risk assessments of real-world applications. According to international standards for information security management such as ISO/IEC 27001, these assessments should  evaluate the likelihood of threats and the potential impact if they materialize. Our code book that evolved based on participants’ statements on attack relevance confirms this approach (Table~\ref{app:tab:codesqRelevance}). In addition, the concrete codes for ‘relevance’ and ‘impact’ that we found in developing our codebook can be used as an orientation for practitioners in trying to concretize likelihood and impact within risk assessments of AI systems. For example, we show that an AI auditor should evaluate concrete implications of ‘financial loss’ or ‘business information leakage' in order to assess the potential impact of an AI risk that might materialize. Thus, our findings allow risk assessments of AI systems to become more granular. 

\myparagraph{Improved communication of AML risks} We provide insights into why practitioners think specific attacks are relevant or irrelevant (Sect.~\ref{sec:prgaRelevance}). These insights into the rationale of relevance for attacks could be a starting point for educational measures to increase AML awareness in organizations that deploy AI. More concretely, our results could help educating business stakeholders that they, for example, have to hedge against model stealing because it is a potential target for IP theft, or that they should consider the tangible risk of poisoning as it may affect their decision-making regarding technology setups or engineering processes. This unveiling of rationales behind attacks might ease the communication of AML risks.

\myparagraph{Open research questions} As we have seen in the previous section, our study comes with some limitations that can be overcome in future work. For example the dimensions knowledge, role in team and the application area deserve to be studied more in depth. Along these lines, we found that not only the application, but also the deployment is a crucial factor determining the vulnerability of an ML system (Sect.~\ref{sec:prgaRelevance}). Both factors need to be monitored and can then jointly, for example together with exposure and AI maturity, be used to assess risks in practice. Such an assessment is also helpful to understand how high the risk of an AML attack is truly—as our 16\% exposure does not take into account cases where an attack would be virtually impossible due to the deployment setting, for example.

Also, some aspects of the relationship between knowledge and concern about threats (Sect.~\ref{sec:practitioners}) remain unclear. This relationship is similar to a chicken-egg problem: more knowledge might imply more sensitivity towards threats, but at the same time more concern also brings about the need for more information, (hopefully) leading to the acquisition of more knowledge. A clear understanding of cause and effect here would benefit regulation and AI in practice.

\section{Related work}
In this section, we first put our findings in relation to other surveys that address AML in practice. Afterwards, we discuss works that are overall less related, yet provide important findings in relation to our insights.

As there is no other survey with a similar amount of participants and thus depth and statistical findings, we instead discuss specific findings from previous works that our larger sample either confirms or contradicts.
For example, Kumar et al.~\cite{kumar2020adversarial} found that poisoning is the most feared AI security threat by companies, a finding we confirm (Fig.~\ref{fig:relevance}). Our findings also support, analogous to their reports, that practitioners are still very much concerned with traditional security (Sect.~\ref{sec:amlQOne})~\cite{kumar2020adversarial}.
 Kumar et al.~\cite{kumar2020adversarial} furthermore state that AML is generally perceived as ``futuristic''. We can not confirm this, however have to emphasize that there are 2-3 years between our surveys. 
Boenisch et al.~\cite{boenisch2021never} find that their ML-security score is overall low, but is increased when participants are developing, and not only applying, ML. We can not confirm these results (Sect.~\ref{sec:practitioners}), however have to emphasize that while we measure direct concern for one ML attack or concern of AML attacks in general, Boenisch et al.'s~\cite{boenisch2021never} score entails also cybersecurity, and ML in general.%
  They~\cite{boenisch2021never} furthermore study privacy, which we only consider in the sense of membership inference. 
Mirsky et al.~\cite{mirsky2021threat} also conducted a survey that aims to understand offensive AI. Specifically for AML, they asked participants how profitable, harmful, detectable and achievable attacks are. We go a step further in our survey and allow our subjects to freely reason why (or why not) they believe an attack to be relevant (Sect.~\ref{sec:prgaRelevance}). Finally, Mirsky et al.~\cite{mirsky2021threat} expect offensive AI techniques to manifest within the next 12 months, a finding our sample does not support (Sect.~\ref{sec:AMLQTwo}).

More loosely related is the work by Bieringer et al.~\cite{bieringer2021mental} who conducted semi-structured interviews with industrial AI practitioners. They find that malicious ML circumventions already take place in industry, which we confirm (Sect.~\ref{sec:AMLQTwo}).  Our study answers their question about the importance of education in the affirmative (Sect.~\ref{sec:practitioners}). 
Finally, we investigated the influence of gender on (ML) security perception, as prior works found that women are overall more concerned about \nonMLSecurity~\cite{hossain2019security}. While we do not find an overall difference, some specific attacks are rated differently (Sect.~\ref{sec:practitioners}), possibly a consequence of women working in slightly different applications~\cite{marini1996gender}.

Finally, our finding of conflated \security and \safety concepts (Sect.~\ref{sec:amlQOne} and Sect.~\ref{sec:prgaRelevance}) has been reported in the cybersecurity domain: For example, Gross and Rosson~\cite{gross2007looking} found in their study that end users do not distinguish system failure from external attacks, and reason that this is a valid level of abstraction for consumers. 
Moreover, early work from 1992 in non-AML security by Loch et al.~\cite{loch1992threats} showed that managers, when introducing information systems to their organizations, ranked intentional \security lower than \safety.
Both works, albeit carried out on different populations than our study, help contextualise our findings (Sect.~\ref{sec:amlQOne} and Sect.~\ref{sec:prgaRelevance}): a similar gap between tomorrow’s reality and today’s understanding might also apply for AML.

\section{Conclusion}
To overcome the lack of knowledge on AML in practice, we conducted a survey of 139 industrial practitioners opinions on ML security and attack relevance.  We found evidence for AML attacks, more specifically evasion and poisoning, in practice. However, it remains often unclear whether an incident is a \security or \safety issue. In addition, also privacy, ML, and organizational challenges like data drift are of importance to our participants. 
In terms of the organizational aspects of AML we find that companies from some industry areas like IT security are more concerned about some ML attacks.
We find no variable that is statistically related to threat exposure. Exposure, together with expected risk and AI maturity, however predicts the amount of implemented mitigations.
We furthermore find that the presence or absence of concern for an AML attack is complex, encompassing factors such as financial loss, ethical concerns, decision making, but also application setting and the way in which ML is deployed.
Finally, on the individual level, we find that self reported knowledge, in particular in AML, increases attack concern. 
Our results yield important insights for regulators and auditors as we analyze relevance and irrelevance, and point out that the boundary between \safety and \security is not always clear. We are further confident that we are contributing towards more research eliciting when ML systems are vulnerable and which factors influence vulnerability and threat perception.

%

\section*{Acknowledgments}
The authors are deeply grateful to all our pre-testers and participants.
We would further like to thank Beat Busser, Federico Marengo, Brian Pendleton, and Jessica Rose for supporting our recruitment. The research reported in this paper has been partly funded by BMK, BMDW, and the Province of Upper Austria in the frame of the COMET Programme managed by FFG in the COMET Module S3AI; and by Fondazione di Sardegna under the project ``TrustML: Towards Machine Learning that Humans Can Trust’’, CUP: F73C22001320007.

\ifCLASSOPTIONcaptionsoff
  \newpage
\fi


\begin{thebibliography}{10}
\providecommand{\url}[1]{#1}
\csname url@samestyle\endcsname
\providecommand{\newblock}{\relax}
\providecommand{\bibinfo}[2]{#2}
\providecommand{\BIBentrySTDinterwordspacing}{\spaceskip=0pt\relax}
\providecommand{\BIBentryALTinterwordstretchfactor}{4}
\providecommand{\BIBentryALTinterwordspacing}{\spaceskip=\fontdimen2\font plus
\BIBentryALTinterwordstretchfactor\fontdimen3\font minus
  \fontdimen4\font\relax}
\providecommand{\BIBforeignlanguage}[2]{{%
\expandafter\ifx\csname l@#1\endcsname\relax
\typeout{** WARNING: IEEEtran.bst: No hyphenation pattern has been}%
\typeout{** loaded for the language `#1'. Using the pattern for}%
\typeout{** the default language instead.}%
\else
\language=\csname l@#1\endcsname
\fi
#2}}
\providecommand{\BIBdecl}{\relax}
\BIBdecl

\bibitem{barreno2006can}
M.~Barreno, B.~Nelson, R.~Sears, A.~D. Joseph, and J.~D. Tygar, ``Can machine
  learning be secure?'' in \emph{CCS}, 2006, pp. 16--25.

\bibitem{biggio2018wild}
B.~Biggio and F.~Roli, ``Wild patterns: Ten years after the rise of adversarial
  machine learning,'' \emph{Patt. Rec.}, vol.~84, pp. 317--331, 2018.

\bibitem{chen2017targeted}
X.~Chen, C.~Liu, B.~Li, K.~Lu, and D.~Song, ``Targeted backdoor attacks on deep
  learning systems using data poisoning,'' \emph{arXiv}, 2017.

\bibitem{cina2022wild}
A.~E. Cin{\`a}, K.~Grosse, A.~Demontis, S.~Vascon, W.~Zellinger, B.~A. Moser,
  A.~Oprea, B.~Biggio, M.~Pelillo, and F.~Roli, ``Wild patterns reloaded: A
  survey of machine learning security against training data poisoning,''
  \emph{arXiv}, 2022.

\bibitem{Dalvi:2004:AC:1014052.1014066}
N.~Dalvi, P.~Domingos, Mausam, S.~Sanghai, and D.~Verma, ``Adversarial
  classification,'' in \emph{KDD}, 2004, pp. 99--108.

\bibitem{gu_badnets_2017}
T.~Gu, B.~Dolan-Gavitt, and S.~Garg, ``Badnets: Identifying vulnerabilities in the ML model supply chain,'' \emph{arXiv}, 2017.

\bibitem{ji2017backdoor}
Y.~Ji, X.~Zhang, and T.~Wang, ``Backdoor attacks against learning systems,'' in \emph{IEEE CNS}, 2017, pp. 1--9.

\bibitem{oh2019towards}
S.~J. Oh, B.~Schiele, and M.~Fritz, ``Towards reverse-engineering black-box
  neural networks,'' in \emph{Explainable AI: Interpreting, Explaining and
  Visualizing Deep Learning}, 2019, pp. 121--144.

\bibitem{papernot2016transferability}
N.~Papernot, P.~McDaniel, and I.~Goodfellow, ``Transferability in machine
  learning: from phenomena to black-box attacks using adversarial samples,''
  \emph{arXiv:1605.07277}, 2016.

\bibitem{DBLP:journals/corr/SzegedyZSBEGF13}
C.~Szegedy, W.~Zaremba, I.~Sutskever, J.~Bruna, D.~Erhan, I.~J. Goodfellow, and
  R.~Fergus, ``Intriguing properties of neural networks,'' in \emph{ICLR},
  2014.

\bibitem{DBLP:conf/uss/TramerZJRR16}
F.~Tram{\`{e}}r, F.~Zhang, A.~Juels, M.~K. Reiter, and T.~Ristenpart,
  ``Stealing machine learning models via prediction APIs,'' in \emph{USENIX Sec.},
  2016, pp. 601--618.

\bibitem{2016arXiv161005820S}
R.~Shokri, M.~Stronati, C.~Song, and V.~Shmatikov, ``Membership inference
  attacks against machine learning models,'' in \emph{S\&P}, 2017, pp. 3--18.

\bibitem{tramer2020adaptive}
F.~Tramer, N.~Carlini, W.~Brendel, and A.~Madry, ``On adaptive attacks to
  adversarial example defenses,'' \emph{NeurIPS}, vol.~33, pp. 1633--1645, 2020.

\bibitem{sommer2010outside}
R.~Sommer and V.~Paxson, ``Outside the closed world: On using machine learning
  for network intrusion detection,'' in \emph{S\&P}, 2010, pp. 305--316.

\bibitem{gilmer2018motivating}
J.~Gilmer, R.~P. Adams, I.~Goodfellow, D.~Andersen, and G.~E. Dahl,
  ``Motivating the rules of the game for adversarial example research,''
  \emph{arXiv}, 2018.

\bibitem{kumar2020adversarial}
R.~S.~S. Kumar, M.~Nystr{\"o}m, J.~Lambert, A.~Marshall, M.~Goertzel,
  A.~Comissoneru, M.~Swann, and S.~Xia, ``Adversarial machine learning-industry
  perspectives,'' in \emph{S\&P Workshops)}, 2020, pp. 69--75.

\bibitem{mirsky2021threat}
Y.~Mirsky, A.~Demontis, J.~Kotak, R.~Shankar, D.~Gelei, L.~Yang, X.~Zhang,
  W.~Lee, Y.~Elovici, and B.~Biggio, ``The threat of offensive ai to
  organizations,'' \emph{arXiv}, 2021.

\bibitem{bieringer2021mental}
L.~Bieringer, K.~Grosse, M.~Backes, B.~Biggio, and K.~Krombholz, ``Industrial
  practitioners' mental models of adversarial machine learning,'' in
  \emph{SOUPS}, 2022, pp. 97--116.

\bibitem{boenisch2021never}
F.~Boenisch, V.~Battis, N.~Buchmann, and M.~Poikela, ``“i never thought about
  securing my machine learning systems”: A study of security and privacy
  awareness of machine learning practitioners,'' in \emph{Mensch und Computer},
  2021, pp. 520--546.

\bibitem{muller2007personal}
P.~J. Muller, S.~E. Young, and M.~N. Vogt, ``Personal rapid transit safety and
  security on university campus,'' \emph{Transportation research record}, vol.
  2006, no.~1, pp. 95--103, 2007.

\bibitem{rubinstein2009antidote}
B.~I. Rubinstein, B.~Nelson, L.~Huang, A.~D. Joseph, S.-h. Lau, S.~Rao,
  N.~Taft, and J.~D. Tygar, ``Antidote: understanding and defending against
  poisoning of anomaly detectors,'' in \emph{IMC}, 2009, pp. 1--14.

\bibitem{biggio2011support}
B.~Biggio, B.~Nelson, and P.~Laskov, ``Support vector machines under
  adversarial label noise,'' in \emph{ACML}, 2011, pp. 97--112.

\bibitem{nasr2018machine}
M.~Nasr, R.~Shokri, and A.~Houmansadr, ``Machine learning with membership
  privacy using adversarial regularization,'' in \emph{CCS}, 2018.

\bibitem{juuti2019prada}
M.~Juuti, S.~Szyller, S.~Marchal, and N.~Asokan, ``Prada: protecting against
  dnn model stealing attacks,'' in \emph{EuroS\&P}, 2019.

\bibitem{lin2021adversarial}
H.-Y. Lin and B.~Biggio, ``Adversarial machine learning: Attacks from laboratories to the real world,'' \emph{IEEE Comp.}, vol.~54, no.~5, pp. 56--60, 2021.

\bibitem{nachar2008mann}
N.~Nachar \emph{et~al.}, ``The Mann-Whitney U: A test for assessing whether two independent samples come from the same distribution,'' \emph{Tutorials in
  quantitative Methods for Psychology}, vol.~4, no.~1, pp. 13--20, 2008.

\bibitem{ferber1952order}
R.~Ferber, ``Order bias in a mail survey,'' \emph{Journal of Marketing},
  vol.~17, no.~2, pp. 171--178, 1952.

\bibitem{huaman2021large}
N.~Huaman, B.~von Skarczinski, C.~Stransky, D.~Wermke, Y.~Acar,
  A.~Drei{\ss}igacker, and S.~Fahl, ``A large-scale interview study on
  information security in and attacks against small and medium-sized
  enterprises,'' in \emph{USENIX Security}, 2021.

\bibitem{mandia2001incident}
K.~Mandia and C.~Prosise, \emph{Incident response: investigating computer
  crime}.\hskip 1em plus 0.5em minus 0.4em\relax McGraw-Hill, Inc., 2001.

\bibitem{european2003commission}
E.~U. Commission \emph{et~al.}, ``Commission recommendation of 6 may 2003
  concerning the definition of micro, small and medium-sized enterprises,''
  \emph{official Journal of the EU}, vol.~46, no. L124, pp. 36--41,
  2003.

\bibitem{samek2017explainable}
W.~Samek, T.~Wiegand, and K.-R. M{\"u}ller, ``Explainable artificial
  intelligence: Understanding, visualizing and interpreting deep learning
  models,'' \emph{arXiv}, 2017.

\bibitem{miller1968response}
R.~B. Miller, ``Response time in man-computer conversational transactions,'' in
  \emph{FJCC}, 1968, pp. 267--277.

\bibitem{suresh2021beyond}
H.~Suresh, S.~R. Gomez, K.~K. Nam, and A.~Satyanarayan, ``Beyond expertise and
  roles: A framework to characterize the stakeholders of interpretable machine
  learning and their needs,'' in \emph{CHI}, 2021.

\bibitem{arrieta2020explainable}
A.~B. Arrieta, N.~D{\'\i}az-Rodr{\'\i}guez, J.~Del~Ser \emph{et~al.}, ``Explainable artificial intelligence (XAI): Concepts,
  taxonomies, opportunities and challenges toward responsible AI,''
  \emph{Information Fusion}, vol.~58, pp. 82--115, 2020.

\bibitem{kaggle}
\BIBentryALTinterwordspacing
Kaggle. (2021) State of machine learning and data science. [Online]. Available:
  \url{https://storage.googleapis.com/kaggle-media/surveys/Kaggle's\%20State\%20of\%20Machine\%20Learning\%20and\%20Data\%20Science\%202021.pdf}
\BIBentrySTDinterwordspacing

\bibitem{strauss1990basics}
A.~Strauss and J.~Corbin, \emph{Basics of qualitative research}.\hskip 1em plus
  0.5em minus 0.4em\relax Sage publications, 1990.

\bibitem{mcdonald2019reliability}
N.~McDonald, S.~Schoenebeck, and A.~Forte, ``Reliability and inter-rater reliability in qualitative research: Norms and guidelines for CSCW and HCI practice,'' \emph{ACM on HCI}, vol.~3, no. CSCW, pp.
  1--23, 2019.

\bibitem{jinyuan2016correlation}
L.~Jinyuan, T.~Wan, C.~Guanqin, L.~Yin, F.~Changyong \emph{et~al.},
  ``Correlation and agreement: overview and clarification of competing concepts
  and measures,'' \emph{Shanghai Archives Psych.}, vol.~28, no.~2, p.
  115, 2016.

\bibitem{cohen1960coefficient}
J.~Cohen, ``A coefficient of agreement for nominal scales,'' \emph{Educational
  and psychological measurement}, vol.~20, no.~1, pp. 37--46, 1960.

\bibitem{hossain2019security}
M.~A. Hossain, ``Security perception in the adoption of mobile payment and the
  moderating effect of gender,'' \emph{PSU Research Review}, 2019.

\bibitem{marini1996gender}
M.~M. Marini, P.-L. Fan, E.~Finley, and A.~M. Beutel, ``Gender and job
  values,'' \emph{Sociology of Education}, pp. 49--65, 1996.

\bibitem{gross2007looking}
J.~B. Gross and M.~B. Rosson, ``Looking for trouble: understanding end-user
  security management,'' in \emph{CHIMIT}, 2007, pp. 10--es.

\bibitem{loch1992threats}
K.~D. Loch, H.~H. Carr, and M.~E. Warkentin, ``Threats to information systems:
  today's reality, yesterday's understanding,'' \emph{Mis Quarterly}, pp.
  173--186, 1992.

\end{thebibliography}

%

\appendices
\section{\textbf{Questionnaire}}\label{app:questionnaire}

\begin{table*}[htbp]
\small
\caption{Codes used to encode the first question, where participants describe their current AI security concerns.}\label{app:tab:codesqOne}
    \centering
    \begin{tabular}{llllllll}
      \toprule
      Group & Code & Group & Code & Group & Code & Group & Code \\
      \midrule
        AML & General & NonAMLSec & General & Privacy & General & ML & General \\   
        ~ & Poisoning & ~ & Libraries & ~ & Regulations & ~ & Explainability \\   
        ~ & Evasion & ~ & Access & & & ~ & Bias \\   
        ~ & Model Stealing & ~ & CustomerIsRisk & Data & General & ~ & Concept drift \\   
        ~ & Performance impact & ~ & CodeBreach & ~ & Data sharing && \\   
        ~ & Robustness & ~ & 3rdParty Provider & ~ & Breach &  Organization & Complexity \\   
        ~ & TestTime & ~ & Precise threat & ~ & Sensitive data & ~ & IP \\  
        ~ & TrainingTime & ~ & Cloud & ~ & Classify if sensitive& ~ & TradeOffs \\  
        ~ & ModelItself & ~ & ~ & ~ & ~ & ~ &SecurityAwareness \\ 
        ~ &  & ~ & ~ & ~ & ~ & ~ & Human Harm \\ 
        \bottomrule
    \end{tabular}
\end{table*}

\begin{table*}[htbp]
\small
    \centering
        \caption{Codes used for the attack relevance, where participants argue why (or why not) an AML attack is relevant or not.}\label{app:tab:codesqRelevance}
    \begin{tabular}{llllll}
    \toprule
    Group & Code & Group & Code & Group & Code \\
    \midrule
        Relevance & General relevance & Impact & General business & Defense & Easy to defend \\  
        ~ & General irrelevance & ~ & Financial loss & ~ & Hard to defend \\  
        ~ & Easy to do & ~ & Business information leakage & ~ & Data access control \\  
        ~ & Hard to do & ~ & Profit for competitor & ~ & Model acccess control \\  
        ~ & Has encountered threat & ~ & Intellectual Property & ~ & No sensitive data \\  
        ~ & Has not encountered threat & ~ & Reputational damage & ~ & Model shortlived \\  
        ~ & Attacker credible & ~ & Regulatory compliance & ~ & Human in the loop \\  
        ~ & Doubting attacker & ~ & Data breach & ~ & Implemented \\  
        ~ & relevant in application setting & ~ & Wrong decision making & & \\  
        ~ & not relevant in use case & ~ & Human harm &  &\\ 
        ~ & not relevant for deployment & ~ & Ethics/Fairness/Bias & Perception & Did not understand threat scenario   \\  
        ~ & Understands attack mechanism & ~ & ~ & ~ & Confusion across threat models \\  
        ~ & Theoretical exposure to threat & ~ & ~ & ~ & Externalization of responsibility \\  
        ~ & Other threat more likely & ~ & ~ & ~ & ~ \\  
        ~ & Safety & ~ & ~ & ~ & ~ \\  
        \bottomrule
    \end{tabular}
\end{table*}

\textbf{Part I - Security of AI within your organization} 
\newline
\textbf{Q1}: In your daily work and your organization's AI workflows, products or systems - what are the most pressing security challenges? [text field] \\
\textbf{Q2}: Did you already experience a circumvention of your AI based workflows, products or systems? [yes/no] \\
IF YES: 
\textbf{Q2.1}: How many circumventions of your AI based workflows, products or systems have you experienced? [1,2,3,4,$>$4] \\
\textbf{Q2.2}: Please describe the most severe circumvention of your AI based workflows, products or systems. [text field] \\
\textbf{Q3}: How high do you estimate the risk of becoming a victim of an attack related to your AI based workflows, products, or systems within the next 12 months? 
[1 (very low) to 5 (very high)] \\
\textbf{Q4}: Which of the following approaches does your organization implement in terms of the security of your AI based workflows, products, or systems? 
[None, Documentation, Guidelines, Mitigations, Fail safe plans, human in the loop, incident response, security testing, other]\\

You will now be confronted with descriptions of specific threats to the security of AI.
Please think about how these threats might take effect in your AI workflows, products, or systems.\\
\textbf{Q5}: Do you consider the following threat scenario relevant in your work? \\
(placeholder for attacks, see below) [very relevant; relevant; not very relevant; irrelevant; I don’t know; I don’t understand threat  scenario]\\
\textbf{Q6}: Why do you think this threat scenario is (placeholder for previous selection)? [text field] \\
These 2 questions are repeated iteratively all attacks:
\begin{enumerate}
\item[a)] \textbf{Q7,8}: Altering training data to harm model performance during deployment. In other words, the model is optimized on tampered training data, which affects the resulting model. 
\item[b)] \textbf{9,10}: Altering test data to harm model performance during deployment. In other words, the trained model is presented with specially crafted inputs that lead to wrong predictions.
\item[c)] \textbf{11,12}: Altering training data so that the model outputs a chosen class whenever a particular pattern is present in the input data. In other words, altering the training data to contain a certain association between a pattern and a label, the resulting model contains a backdoor.
\item[d)] \textbf{13,14-Sanity}:  Altering training data to delete an untrained model. In other words, the training data contains a pattern that will delete the model after training.
\item[e)] \textbf{15,16}: Given input data and the predictions of a model, determine whether the given data sample is part of the training data. In other words, the model is queried to obtain crucial information about the used training data. 
\item[f)] \textbf{17,18}: Given an API / black box access to a model, copy its functionality. In other words, repeatedly observe input and output pairs from the model to reproduce its functionality. 
\end{enumerate}
\textbf{Part II - AI within your organization}
\newline
\textbf{Q17}: In which country is your organization headquartered?
[drop down with all countries] \\
\textbf{Q18}: What is the number of employees at your organization? [$<$10, 10-49, 50–99, 100–249, 250–499, $>$500] \\
\textbf{Q19}: Which industry area describes your organizations best? [Customer Service \& Support, IT Security, Production, Marketing, Computer Audition, Research, Forecasting, Computer Linguistics, Computer Vision, Agriculture Forestry \& Fishing, Finance \& Insurance, Arts Entertainment \& Recreation, Manufacturing, Water \& Waste, Healthcare, Retail \& Commerce, Transportation \& Mobility, Other] \\
\textbf{Q20}: What kind of data analysis do you work with primarily? [supervised learning, unsupervised learning, semi-supervised learning, reinforcement learning, other] \\
\textbf{Q21}: What do you use AI for primarily (e.g. sentiment analysis, object detection, malware classification)? [text field] \\
\textbf{Q22}: What input data do you work with primarily? (tick most specific) [Images,  Videos, Speech/Audio, Text/Documents, Network traffic, Social media data, Files/Source Code, Other:] \\
\textbf{Q23}: What kind of labels do you work with primarily?  
[unlabelled, categorical, real valued, structured data, other] \\ 
\textbf{Q24}: What is the status of the ML projects you work on?
\begin{itemize}
      \item[]      Indirect usage (e.g. certification, auditing)
\item[] Evaluating use cases
\item[] Starting to develop models
\item[] Getting developed models into production
\item[] Models in production, for 1-2 years
\item[] Models in production, for 2-4 years
\item[] Models in production, for $>$5 years
\end{itemize}
\textbf{Q25}: Which of these goals are part of your organization's ML-model checklist? [Performance, fairness, explainability, security, privacy, ethics, system response, other] \\
\textbf{Part III - Demographics and your AI background}
\newline
\textbf{Q26}: In which year were you born? [1935-2021]\\
\textbf{Q27}: What gender do you identify with?  
[Female,  male,  other,  I do not want to disclose] \\
\textbf{Q28}: In which country are you located? [drop down with all countries]\\
\textbf{Q29}: What is your level of education? Please specify the highest. [Highschool, Bachelor, Master/Diploma, Training/Apprenticeship, PhD, Other] \\
\textbf{Q30}: What is your role in your team? [ML Engineer, ML researcher, Data scientist, Domain Expert, Product Owner, Auditor, Other] \\
\textbf{Q31}: Please complete the following sentence. When it comes to machine learning, I believe I have… [No knowledge, a little/some/moderate/high knowledge] \\
\textbf{Q32}: In which of these areas have you taken a lecture or intense course? 
[None, Machine learning, Security, Adversarial Machine Learning]\\

\section{Detailed comparison with the Kaggle sample}\label{app:kaggcomp}
As we write in Sect.~\ref{sec:sampledescr} in the sample description, our sample matches roughly the numbers from the Kaggle report~\cite{kaggle}, modulo that the report has a larger sample (>25,000 participants) and has made slightly different design choices concerning the questions. In this Appendix, we redraw and reordered the plots from the report to be able to roughly compare them to our data in table~\ref{tab:kaggleComparison}.  
\begin{table*}[]
    \centering
        \caption{Detailed comparison with the Kaggle report~\cite{kaggle}. We reordered the original information to enable an easier comparison. The AI maturity data is from the Kaggle report from 2020, as the report from 2021 does not contain this information. \textbf{Units of plots do not match and have not been adjusted.}}
    \label{tab:kaggleComparison}
    \begin{tabular}{lrr}
    \toprule
        & Our sample  & Kaggle~\cite{kaggle} \\
        \midrule
        \multirow{5}{*}{\rotvertical{\textbf{Gender}}} & 70.1\% male & 82.2\% male \\
        & 14.3\% female & 16.2\% female \\
        && 0.2\% non-binary \\
        & 7\% do not wish to disclose & 1.2\% prefer not to say \\
        & 9\% no reply & 0.2\% prefer to self describe \\
        \rotvertical{\hspace*{20mm}\textbf{Age}} & \includegraphics[  width=.33\textwidth]{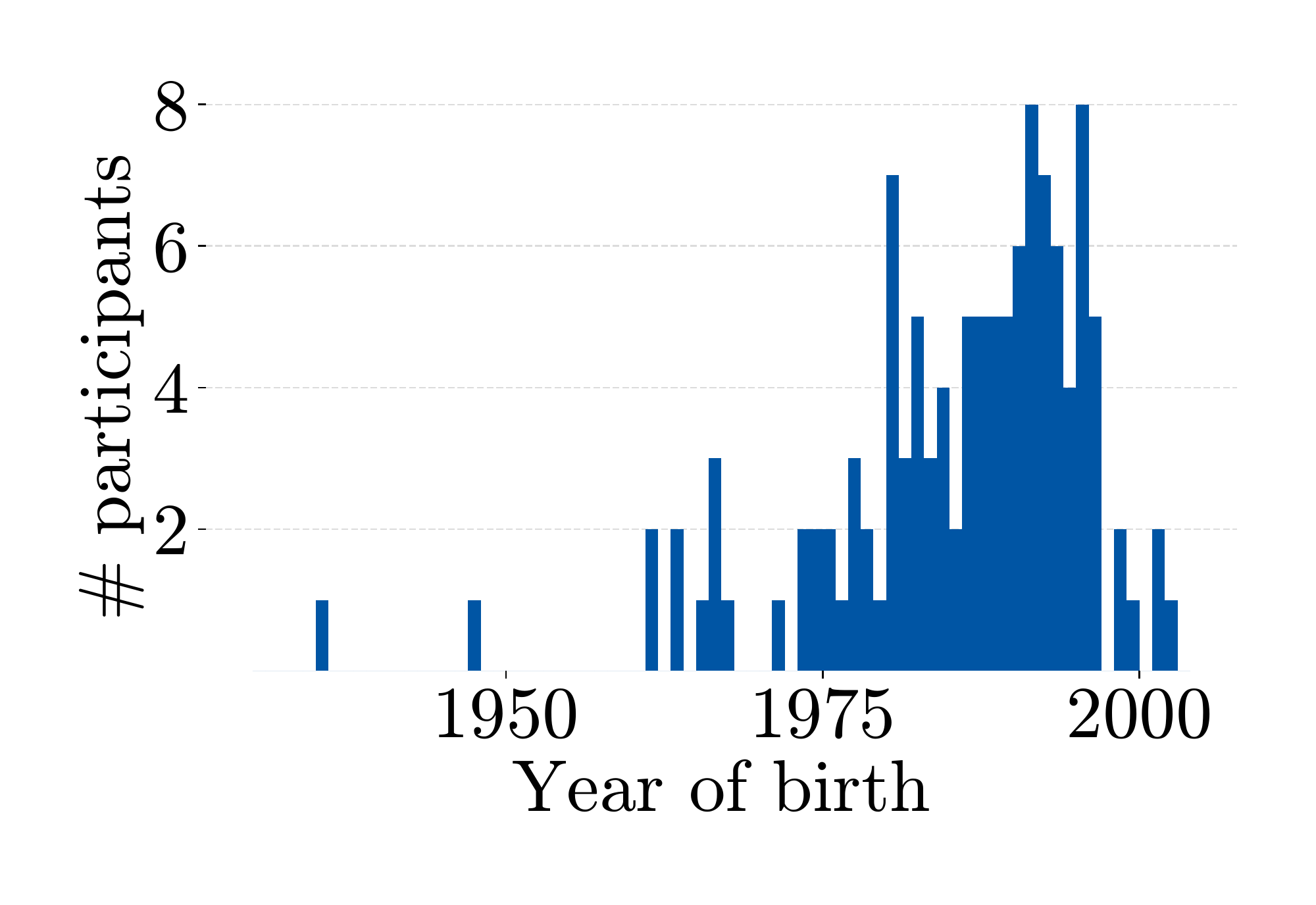} & \includegraphics[  width=.33\textwidth]{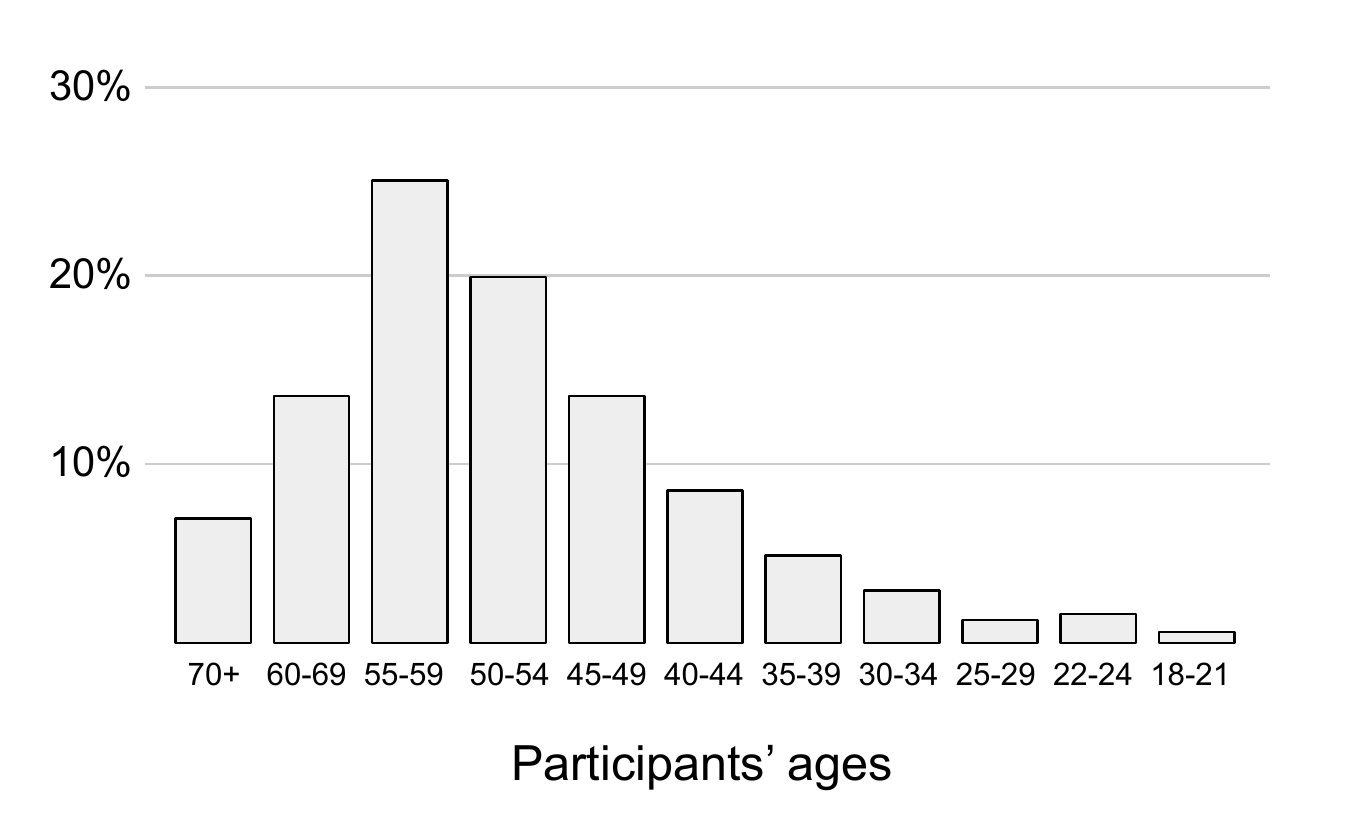} \\
        \rotvertical{\hspace*{15mm}\textbf{Education}} & \includegraphics[  width=.33\textwidth]{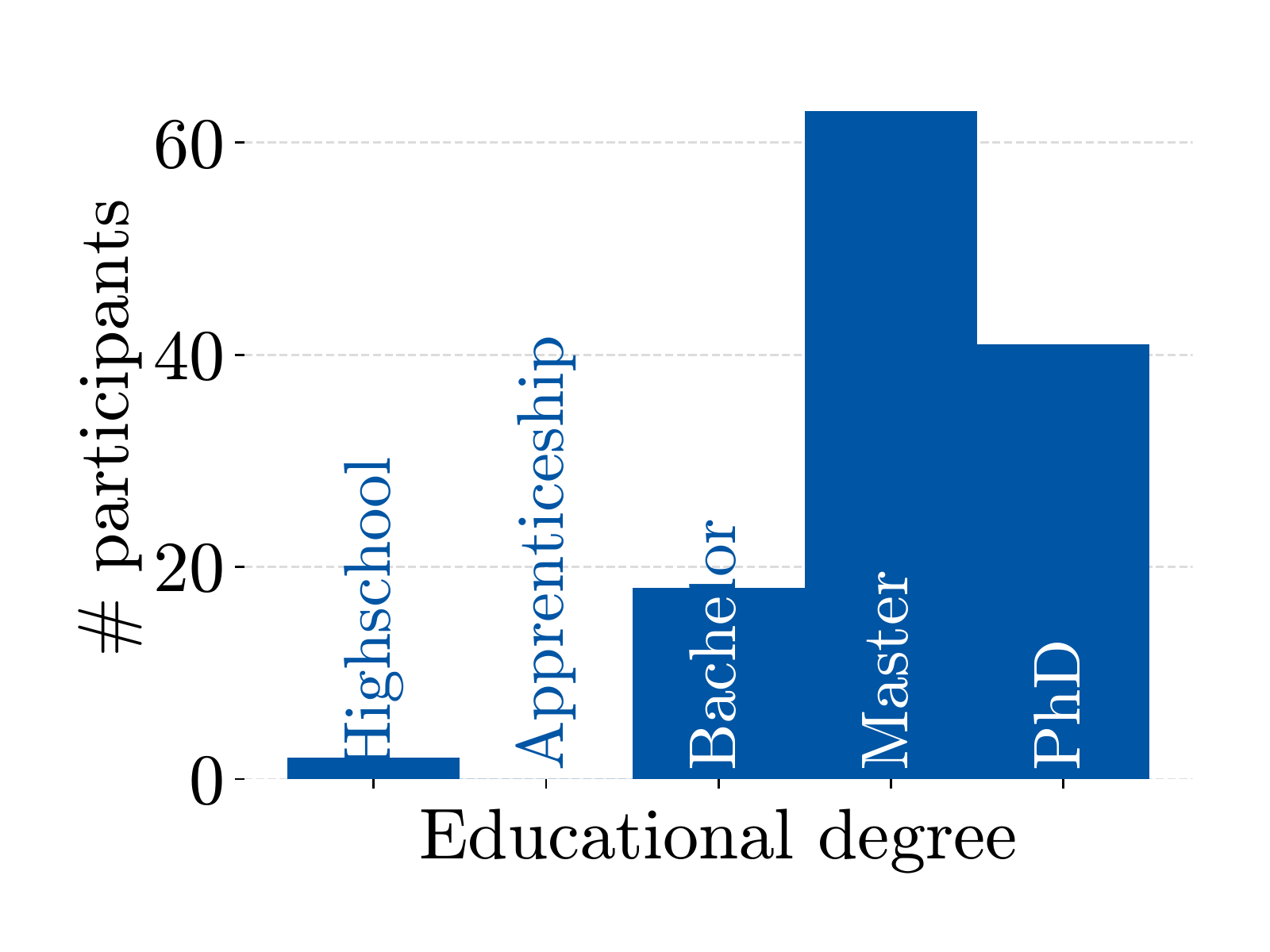} & \includegraphics[  width=.33\textwidth]{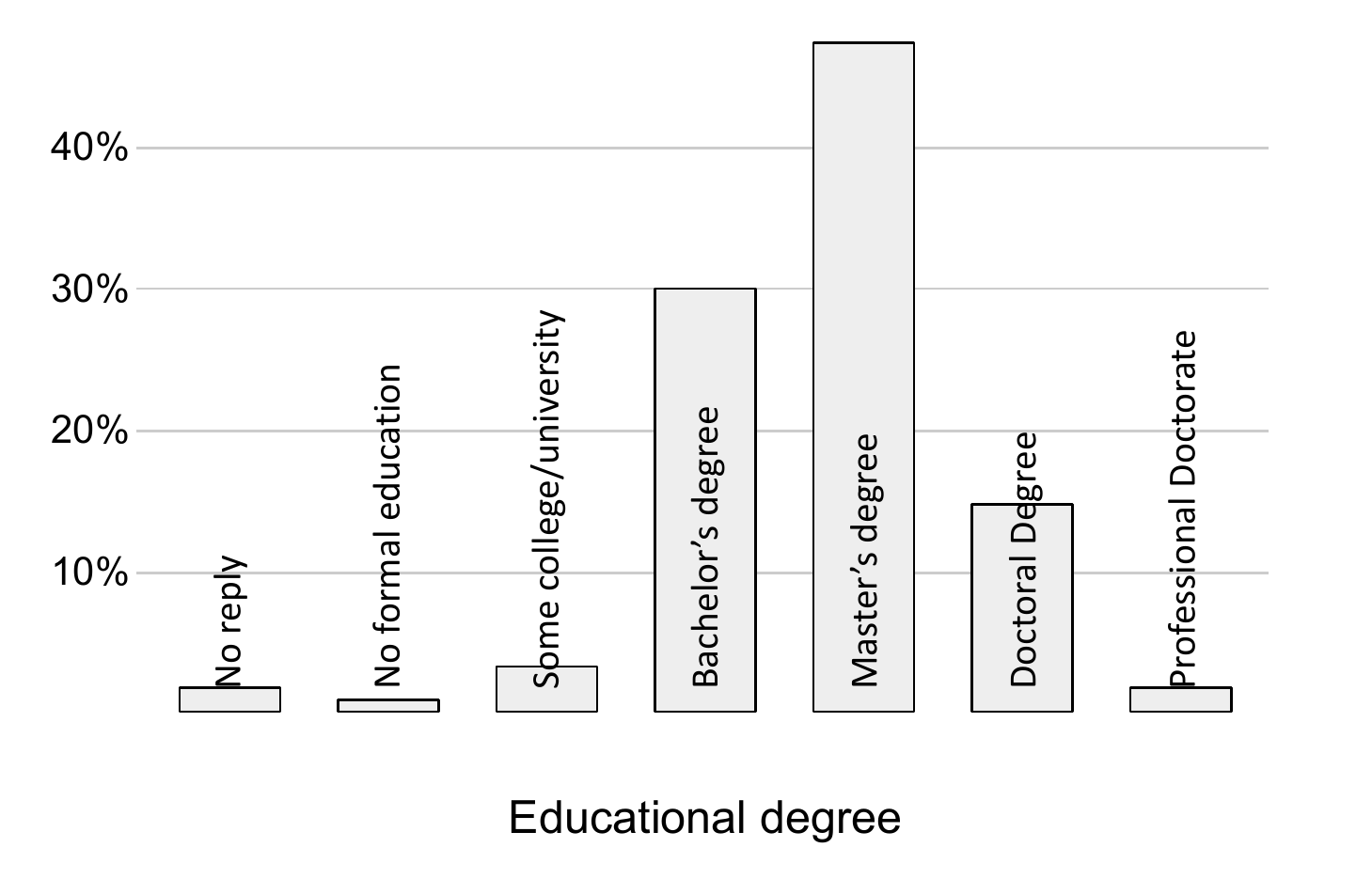} \\
        \rotvertical{\hspace*{13mm}\textbf{Organizations size}} & \includegraphics[  width=.33\textwidth]{figures/CompanySize.pdf} & \includegraphics[  width=.33\textwidth]{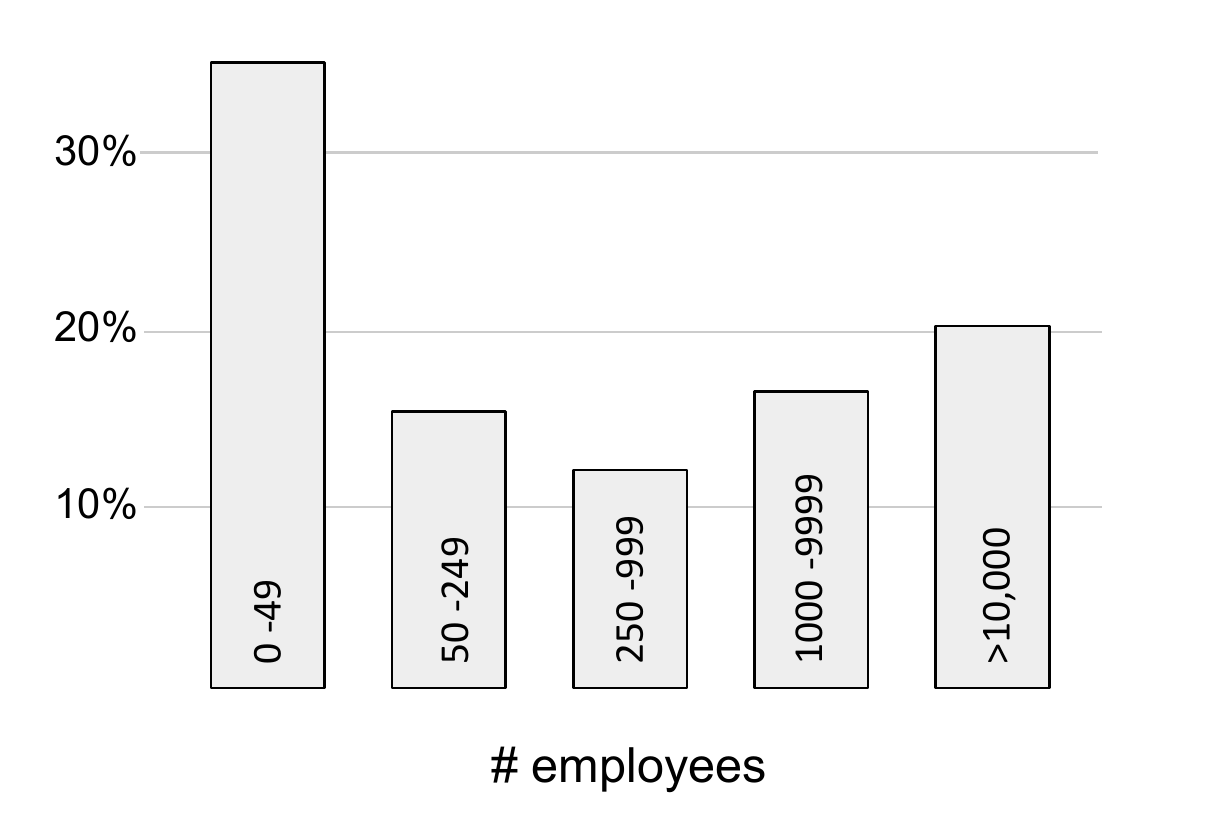} \\
        \rotvertical{\hspace*{15mm}\textbf{AI maturity}} & \includegraphics[  width=.33\textwidth]{figures/Maturity.pdf} & \includegraphics[  width=.33\textwidth]{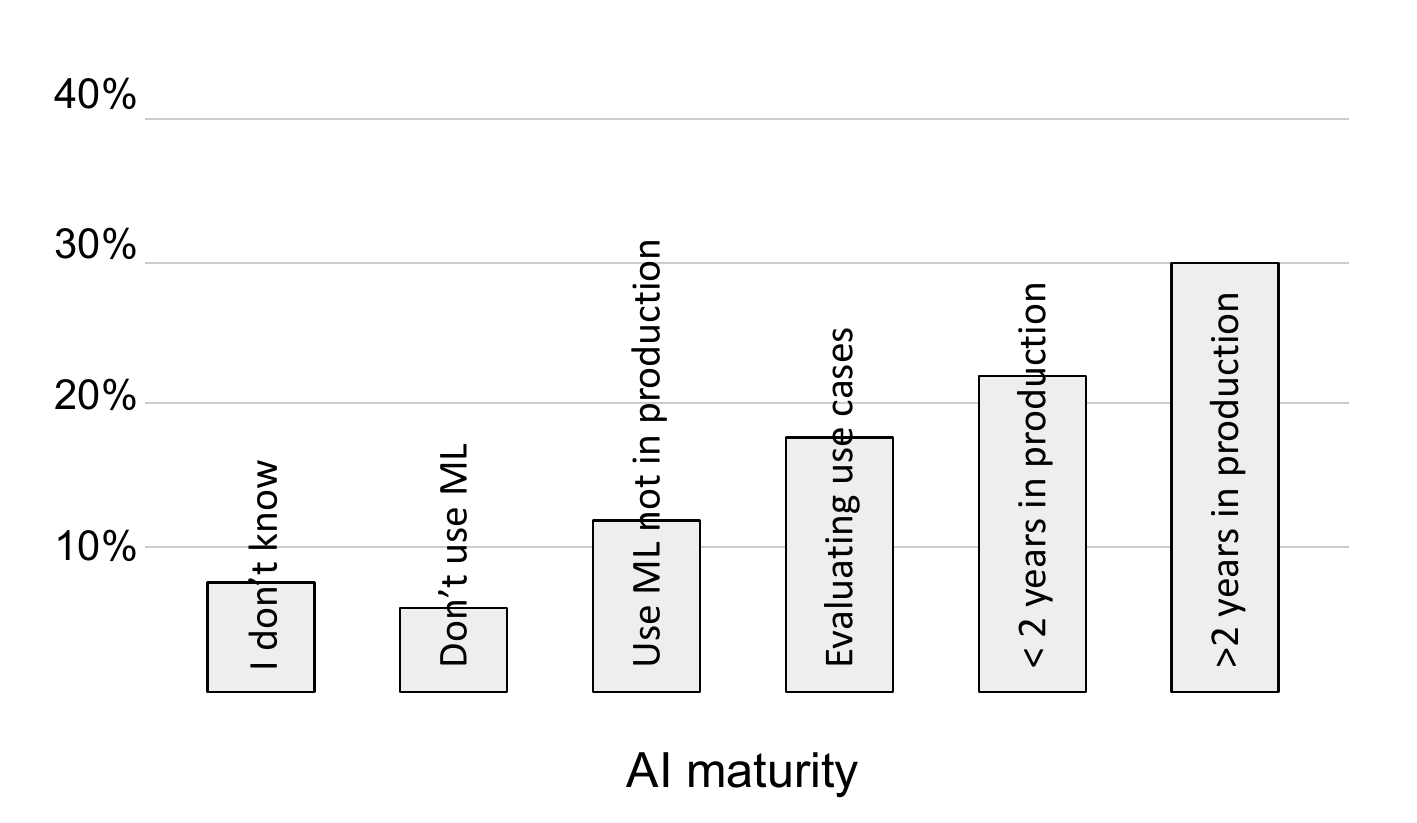} \\
    \bottomrule
    \end{tabular}
\end{table*}

\section{Complete sets of Codes}\label{app:codes}
We here depict the full sets of codes for most feared threat/the first question in Table~\ref{app:tab:codesqOne}. The codes for the attack relevance coding are listed in Table~\ref{app:tab:codesqRelevance}.

\section{Detailed results of statistical tests}\label{app:stattests}

We here report the detailed results for all statistical tests in the main paper in the order of appearance or mentioning.

\subsection{Statistical tests from Section~\ref{sec:orgaStatTests}}
We first review the tests from Section~\ref{sec:orgaStatTests} about the participant's organization in the order of the paragraphs in the main paper.

\myparagraph{Organization Area and threat perception} In this part, we depict the detailed results of the organization area and threat relevance ratings. The table takes organizational area (\question{19}), and splits according to Healthcare / Security the provided rating (e.g., participants working in a healthcare setting vs all other participants). We depict mean ($\pm$ standard deviation) and the sample size for each subgroup (left/sample I: works in specified industry, right/sample II: remaining participants) as well as the test statistic and the $p$-value.
\\\begin{tabular}{lrrrrr} \\ 
 \toprule 
 \multicolumn{2}{l}{Sample I} & \multicolumn{2}{l}{Sample II} \\
 \cmidrule(r){1-2} \cmidrule(r){3-4}
 $\mu$ ($\pm $sdt.) & \# &  $\mu$ ($\pm $sdt.) & \# & U & $p$ \\
 \midrule
\multicolumn{4}{l}{Healthcare: Membership Rating (\question{13})} \\ 
1.67 ($\pm$ 1.49) & 18 & 1.71 ($\pm$ 1.85) & 121 & 1030.0 & 0.71 \\ 
 \addlinespace \multicolumn{4}{l}{Security: Poisoning Rating (\question{5})} \\ 
3.0 ($\pm$ 0.97) & 15 & 2.3 ($\pm$ 1.38) & 124 & 1200.0 & 0.06 \\ 
 \addlinespace \multicolumn{4}{l}{Security: Evasion Rating (\question{7})} \\ 
2.73 ($\pm$ 1.44) & 15 & 1.9 ($\pm$ 1.61) & 124 & 1262.0 & \textbf{0.02} \\ 
 \addlinespace \multicolumn{4}{l}{Security: Backdoor Rating (\question{9})} \\ 
3.0 ($\pm$ 0.97) & 15 & 1.6 ($\pm$ 1.74) & 124 & 1372.5 & \textbf{0.0} \\ 
 \addlinespace \multicolumn{4}{l}{Security: Sanity Rating (\question{11})} \\ 
0.4 ($\pm$ 2.06) & 15 & 0.24 ($\pm$ 1.82) & 124 & 953.0 & 0.88 \\ 
 \addlinespace \multicolumn{4}{l}{Security: Membership Rating (\question{13})} \\ 
2.6 ($\pm$ 1.5) & 15 & 1.6 ($\pm$ 1.81) & 124 & 1251.5 & \textbf{0.03} \\ 
 \addlinespace \multicolumn{4}{l}{Security: M. Stealing Rating (\question{15})} \\ 
2.13 ($\pm$ 1.5) & 15 & 1.65 ($\pm$ 1.81) & 124 & 1047.0 & 0.42 \\ 
\bottomrule \end{tabular}

 \myparagraph{Predicting threat exposure} Analogous to the previous tests, we again divide according to company area (\question{19}) and now test for exposure.
  \vspace{0.5em}
 \begin{tabular}{lrrrrr} \\ 
 \toprule
  \multicolumn{2}{l}{Sample I} & \multicolumn{2}{l}{Sample II} \\
 \cmidrule(r){1-2} \cmidrule(r){3-4}
 $\mu$ ($\pm $sdt.) & \# &  $\mu$ ($\pm $sdt.) & \# & U & $p$ \\
 \midrule
\multicolumn{4}{l}{Security: exposure (\question{2})} \\ 
0.44 ($\pm$ 1.21) & 18 & 0.51 ($\pm$ 1.32) & 121 & 1079.0 & 0.93 \\ 
 \addlinespace \multicolumn{4}{l}{Healthcare: exposure (\question{2})} \\ 
1.0 ($\pm$ 1.86) & 15 & 0.44 ($\pm$ 1.21) & 124 & 1038.5 & 0.27 \\ 
 \bottomrule \end{tabular}
  \vspace{0.5em}
  
 We furthermore ran a regression model optimized with bfgs, with the predictors AI maturity (\question{24}) and company size (\question{18}). The obtained log-likelihood of the model was $-96.907$, the AIC $207.8$ and the BIC $228.4$. The model had 132 residuals and 7 degrees of freedom.
 
 \vspace{0.5em}
 
\hspace{-0.5em}
\resizebox{0.49\textwidth}{!}{%
\begin{tabular}{@{}lcccccc@{}}
\toprule
                 & coef & std err & z & P$> |$z$|$ & [0.025 & 0.975]  \\
\midrule
AI maturity      &       0.1  &        0.074     &     1.33  &         0.19        &       -0.05    &        0.24     \\
Comp. size      &       0.01  &        0.061     &     0.13  &         0.9        &       -0.11    &        0.13     \\
0.0/1.0 &       1.36  &        0.374     &     3.64  &         0.0       &        0.63    &        2.09     \\
1.0/2.0 &      -1.5  &        0.364     &    -4.11  &         0.0        &       -2.21    &       -0.78     \\
2.0/3.0 &      -1.38  &        0.393     &    -3.51  &         0.0        &       -2.15    &       -0.61     \\
3.0/4.0 &      -2.97  &        0.991     &    -3  &         0.0        &       -4.91    &       -1.03     \\
4.0/5.0 &      -2.15  &        0.695     &    -3.1  &         0.0        &       -3.52    &       -0.79     \\
\bottomrule
\end{tabular}}

\myparagraph{Predicting amount of implemented mitigations}
 To gain insights on the amount of implemented mitigations, we ran a regression model optimized with bfgs. The predictors were AI maturity (\question{24}), company size (\question{18}), exposure (\question{2}) and estimated risk of exposure (\question{3}). As predicted variable, we determine the number of implemented mitigations (min. 0, maximum 7). The obtained log-likelihood was $-266.99$, the AIC $556$ and the BIC $588.3$. The model had 128 residuals and 11 degrees of freedom.

\hspace{-0.7em}
\resizebox{0.49\textwidth}{!}{%
\begin{tabular}{@{}lcccccc@{}}
\toprule
    & coef & std err & z & P$> |$z$|$ & [0.025 & 0.975]  \\
\midrule
AI maturity      &       0.15  &        0.051     &     2.84  &         \textbf{0.00}        &        0.045    &        0.246     \\
Comp. size      &       0.01  &        0.043     &     0.17  &         0.87        &       -0.08    &        0.09     \\
Exposure      &       0.18  &        0.071     &     2.5  &         \textbf{0.01}        &        0.04    &        0.32     \\
Est. risk      &       0.2  &        0.079     &     2.5  &         \textbf{0.01}        &        0.04    &        0.35     \\
0.0/1.0 &      -0.06  &        0.316     &    -0.18  &         0.86        &       -0.68    &        0.56     \\
1.0/2.0 &      -0.57  &        0.211     &    -2.7  &         0.01        &       -0.98    &       -0.15     \\
2.0/3.0 &      -0.57  &        0.179     &    -3.2  &         0.00        &       -0.92    &       -0.22     \\
3.0/4.0 &      -0.68  &        0.180     &    -3.8  &         0.00        &       -1.03    &       -0.32     \\
4.0/5.0 &      -0.91  &        0.220     &    -4.1  &         0.00        &       -1.34    &       -0.47     \\
5.0/6.0 &      -1.02  &        0.273     &    -3.7  &         0.00        &       -1.56    &       -0.49     \\
6.0/7.0 &      -1.24  &        0.362     &    -3.4  &         0.00        &       -1.95    &       -0.53     \\
\bottomrule
\end{tabular}}

 \myparagraph{Comparing attack ratings} As reported in the main paper, the Sanity threat (see App.~\ref{app:questionnaire}) is rated statistically significantly different than the other attacks. We here report the detailed results from the attack vs attack ratings. Here, we use the two samples with the numeric inputs for one attack as one input sample. The statistics ($\mu$,std., number of samples) are reported for each attack. we encode  replies without rating as well. 
 
 In addition to the statistical significance for sanity, we also observe statistical significance for poisoning. We do not report this, as we were not able to randomize the order of the attacks, and we assume this is an effect of our participants getting tired. An alternative explanation is that as reported by Kumar et al.~\cite{kumar2020adversarial}, poisoning is indeed the most feared threat (and thus rated higher then other attacks). In addition, the test statistic is also higher when testing against santity for any attack, showing that this effect is stronger.
  \vspace{0.5em}
  
 \hspace{-0.5em}
 \begin{tabular}{lrrrrr} 
 \toprule
  \multicolumn{2}{l}{Attack I} & \multicolumn{2}{l}{Attack II} \\
 \cmidrule(r){1-2} \cmidrule(r){3-4}
 $\mu$ ($\pm $sdt.) & \# &  $\mu$ ($\pm $sdt.) & \# & U & $p$ \\
 \midrule
\multicolumn{2}{l}{Poisoning  (\question{5})} & \multicolumn{2}{l}{Evasion  (\question{7})} \\ 
2.37 ($\pm$ 1.36) & 139 & 1.99 ($\pm$ 1.61) & 139 & 10908 & 0.05 \\ 
 \addlinespace \multicolumn{2}{l}{Poisoning  (\question{5})} & \multicolumn{2}{l}{Backdoor  (\question{9})} \\ 
2.37 ($\pm$ 1.36) & 139 & 1.76 ($\pm$ 1.73) & 139 & 11557 & \textbf{0.0} \\ 
 \addlinespace \multicolumn{2}{l}{Poisoning  (\question{5})} & \multicolumn{2}{l}{Sanity  (\question{11})} \\ 
2.37 ($\pm$ 1.36) & 139 & 0.26 ($\pm$ 1.85) & 139 & 15796 & \textbf{0.0} \\ 
 \addlinespace \multicolumn{2}{l}{Poisoning  (\question{5})} & \multicolumn{2}{l}{Membership  (\question{13})} \\ 
2.37 ($\pm$ 1.36) & 139 & 1.71 ($\pm$ 1.81) & 139 & 11622 & \textbf{0.0} \\ 
 \addlinespace \multicolumn{2}{l}{Poisoning  (\question{5})} & \multicolumn{2}{l}{M. Stealing  (\question{15})} \\ 
2.37 ($\pm$ 1.36) & 139 & 1.71 ($\pm$ 1.79) & 139 & 11566 & \textbf{0.0} \\ 
 \addlinespace \multicolumn{2}{l}{Evasion  (\question{7})} & \multicolumn{2}{l}{Backdoor  (\question{9})} \\ 
1.99 ($\pm$ 1.61) & 139 & 1.76 ($\pm$ 1.73) & 139 & 10327 & 0.31 \\ 
 \addlinespace \multicolumn{2}{l}{Evasion  (\question{7})} & \multicolumn{2}{l}{Sanity  (\question{11})} \\ 
1.99 ($\pm$ 1.61) & 139 & 0.26 ($\pm$ 1.85) & 139 & 14669 & \textbf{0.0} \\ 
 \addlinespace \multicolumn{2}{l}{Evasion  (\question{7})} & \multicolumn{2}{l}{Membership  (\question{13})} \\ 
1.99 ($\pm$ 1.61) & 139 & 1.71 ($\pm$ 1.81) & 139 & 10441 & 0.23 \\ 
 \addlinespace \multicolumn{2}{l}{Evasion  (\question{7})} & \multicolumn{2}{l}{M. Stealing  (\question{15})} \\ 
1.99 ($\pm$ 1.61) & 139 & 1.71 ($\pm$ 1.79) & 139 & 10383 & 0.27 \\ 
 \addlinespace \multicolumn{2}{l}{Backdoor  (\question{9})} & \multicolumn{2}{l}{Sanity  (\question{11})} \\ 
1.76 ($\pm$ 1.73) & 139 & 0.26 ($\pm$ 1.85) & 139 & 14065 & \textbf{0.0} \\ 
 \addlinespace \multicolumn{2}{l}{Backdoor  (\question{9})} & \multicolumn{2}{l}{Membership  (\question{13})} \\ 
1.76 ($\pm$ 1.73) & 139 & 1.71 ($\pm$ 1.81) & 139 & 9789 & 0.84 \\ 
 \addlinespace \multicolumn{2}{l}{Backdoor  (\question{9})} & \multicolumn{2}{l}{M. Stealing  (\question{15})} \\ 
1.76 ($\pm$ 1.73) & 139 & 1.71 ($\pm$ 1.79) & 139 & 9728 & 0.92 \\ 
 \addlinespace \multicolumn{2}{l}{Sanity  (\question{11})} & \multicolumn{2}{l}{Membership  (\question{13})} \\ 
0.26 ($\pm$ 1.85) & 139 & 1.71 ($\pm$ 1.81) & 139 & 5551 & \textbf{0.0} \\ 
 \addlinespace \multicolumn{2}{l}{Sanity  (\question{11})} & \multicolumn{2}{l}{M. Stealing  (\question{15})} \\ 
0.26 ($\pm$ 1.85) & 139 & 1.71 ($\pm$ 1.79) & 139 & 5420 & \textbf{0.0} \\ 
 \addlinespace \multicolumn{2}{l}{Membership  (\question{13})} & \multicolumn{2}{l}{M. Stealing  (\question{15})} \\ 
1.71 ($\pm$ 1.81) & 139 & 1.71 ($\pm$ 1.79) & 139 & 9616 & 0.95 \\ 
\bottomrule 
\end{tabular}

\subsection{Statistical tests from Section~\ref{sec:practitioners}}
We first consider the results from Section~\ref{sec:practitioners} about our participants  as in the order of the paragraphs in the main paper.

\textbf{Role in team}
We now split our data according to whether a participant has a technical role (left side) or not (right side).  
\begin{tabular}{lrrrrr} \\
 \toprule 
   \multicolumn{2}{l}{Non-technical} & \multicolumn{2}{l}{Technical} \\
 \cmidrule(r){1-2} \cmidrule(r){3-4}
 $\mu$ ($\pm $sdt.) & \# &  $\mu$ ($\pm $sdt.) & \# & U & $p$ \\
 \midrule
\multicolumn{4}{l}{Exposure (\question{2})} \\ 
0.34 ($\pm$ 1.02) & 91 & 0.81 ($\pm$ 1.68) & 48 & 1970 & 0.15 \\ 
 \addlinespace \multicolumn{4}{l}{Poisoning Rating (\question{5})} \\ 
2.46 ($\pm$ 1.26) & 91 & 2.21 ($\pm$ 1.51) & 48 & 2370 & 0.39 \\ 
 \addlinespace \multicolumn{4}{l}{Evasion Rating (\question{7})} \\ 
1.92 ($\pm$ 1.7) & 91 & 2.12 ($\pm$ 1.41) & 48 & 2135 & 0.83 \\ 
 \addlinespace \multicolumn{4}{l}{Backdoor Rating (\question{9})} \\ 
1.79 ($\pm$ 1.7) & 91 & 1.69 ($\pm$ 1.77) & 48 & 2290 & 0.63 \\ 
 \addlinespace \multicolumn{4}{l}{Sanity Rating (\question{11})} \\ 
0.03 ($\pm$ 1.78) & 91 & 0.69 ($\pm$ 1.92) & 48 & 1756 & 0.05 \\ 
 \addlinespace \multicolumn{4}{l}{Membership Rating (\question{13})} \\ 
1.73 ($\pm$ 1.74) & 91 & 1.67 ($\pm$ 1.93) & 48 & 2151 & 0.89 \\ 
 \addlinespace \multicolumn{4}{l}{M. Stealing Rating (\question{15})} \\ 
1.79 ($\pm$ 1.78) & 91 & 1.54 ($\pm$ 1.79) & 48 & 2340 & 0.48 \\ 
 \bottomrule
 \end{tabular}
 \vspace{0.7em}

 \myparagraph{Prior Knowledge in AML}
 We now divide our sample along the self-reported knowledge of AML (present, Sample I, not present: Sample II). For the security approaches(\question{4}), we encode the replies using the amount of implemented approaches (e.g., 0-7).
 
  \vspace{0.7em}
  
 \begin{tabular}{rlllll}
 \toprule
\multicolumn{2}{l}{Sample I} & \multicolumn{2}{l}{Sample II} \\
 \cmidrule(r){1-2} \cmidrule(r){3-4}
 $\mu$ ($\pm $sdt.) & \# &  $\mu$ ($\pm $sdt.) & \# & U & $p$ \\
 \midrule
\multicolumn{4}{l}{Exposure (\question{2})} \\ 
0.52 ($\pm$ 1.3) & 40 & 0.49 ($\pm$ 1.31) & 99 & 1991 & 0.94 \\ 
 \addlinespace \multicolumn{4}{l}{Security approaches (\question{4})} \\ 
3.45 ($\pm$ 2.02) & 40 & 2.72 ($\pm$ 2.05) & 99 & 2403 & \textbf{0.05} \\ 
 \addlinespace \multicolumn{4}{l}{Estimated Risk (\question{3})} \\ 
5.5 ($\pm$ 2.26) & 40 & 4.93 ($\pm$ 2.35) & 99 & 2227 & 0.25 \\ 
 \addlinespace \multicolumn{4}{l}{Poisoning Rating (\question{5})} \\ 
2.88 ($\pm$ 0.95) & 40 & 2.17 ($\pm$ 1.44) & 99 & 2533 & \textbf{0.01} \\ 
 \addlinespace \multicolumn{4}{l}{Evasion Rating (\question{7})} \\ 
2.65 ($\pm$ 1.22) & 40 & 1.73 ($\pm$ 1.67) & 99 & 2580 & \textbf{0.0} \\ 
 \addlinespace \multicolumn{4}{l}{Backdoor Rating (\question{9})} \\ 
2.38 ($\pm$ 1.58) & 40 & 1.51 ($\pm$ 1.72) & 99 & 2567 & \textbf{0.01} \\ 
 \addlinespace \multicolumn{4}{l}{Sanity Rating (\question{11})} \\ 
0.52 ($\pm$ 1.96) & 40 & 0.15 ($\pm$ 1.79) & 99 & 2183 & 0.33 \\ 
 \addlinespace \multicolumn{4}{l}{Membership Rating (\question{13})} \\ 
2.4 ($\pm$ 1.51) & 40 & 1.42 ($\pm$ 1.84) & 99 & 2581 & \textbf{0.0} \\ 
 \addlinespace \multicolumn{4}{l}{M. Stealing Rating (\question{15})} \\ 
2.35 ($\pm$ 1.28) & 40 & 1.44 ($\pm$ 1.89) & 99 & 2460 & \textbf{0.02} \\ 
 \bottomrule \end{tabular}
 
\myparagraph{ML knowledge and education}
 We now divide our sample along the self-reported knowledge of ML (present, Sample I, not present: Sample II). For the security approaches(\question{4}), we encode the replies using the amount of implemented approaches (e.g., 0-7).
 \begin{tabular}{lrrrrr} \\ 
 \toprule 
 \multicolumn{2}{l}{Sample I} & \multicolumn{2}{l}{Sample II} \\
 \cmidrule(r){1-2} \cmidrule(r){3-4}
 $\mu$ ($\pm $sdt.) & \# &  $\mu$ ($\pm $sdt.) & \# & U & $p$ \\
 \midrule
\multicolumn{4}{l}{Poisoning Relevance (\question{5})} \\ 
2.55 ($\pm$ 1.08) & 114 & 1.56 ($\pm$ 2.04) & 25 & 1738.0 & 0.07 \\ 
 \addlinespace \multicolumn{4}{l}{Evasion Relevance (\question{7})} \\ 
2.16 ($\pm$ 1.49) & 114 & 1.24 ($\pm$ 1.9) & 25 & 1811.0 & \textbf{0.03} \\ 
 \addlinespace \multicolumn{4}{l}{Backdoor Relevance (\question{9})} \\ 
1.88 ($\pm$ 1.64) & 114 & 1.2 ($\pm$ 1.96) & 25 & 1685.0 & 0.15 \\ 
 \addlinespace \multicolumn{4}{l}{Sanity Relevance (\question{11})} \\ 
0.27 ($\pm$ 1.83) & 114 & 0.2 ($\pm$ 1.96) & 25 & 1480.0 & 0.76 \\ 
 \addlinespace \multicolumn{4}{l}{Membership Relevance (\question{13})} \\ 
1.87 ($\pm$ 1.74) & 114 & 0.96 ($\pm$ 1.93) & 25 & 1811.5 & \textbf{0.03} \\ 
 \addlinespace \multicolumn{4}{l}{Model Stealing Relevance (\question{15})} \\ 
1.85 ($\pm$ 1.71) & 114 & 1.04 ($\pm$ 1.97) & 25 & 1769.5 & 0.05 \\ 
\bottomrule \end{tabular}

 To gain insights on the influence of education, we ran a regression model optimized with bfgs. The predictor was education (\question{29}), the predicted variable exposure (\question{2}). The obtained log-likelihood was $-96.77$, the AIC $205.5$ and the BIC $223.1$. The model had 133 residuals and 6 degrees of freedom.
 
 \vspace{0.5em}
 
 \hspace{-0.7em}
\resizebox{0.49\textwidth}{!}{%
\begin{tabular}{lcccccc}
\toprule
                 & coef & std err & z & P$> |$z$|$ & [0.025 & 0.975]  \\
\midrule
Education  &    0.133  &        0.096     &     1.38  &         0.17        &       -0.06    &        0.32     \\
0.0/1.0 &       1.447  &        0.395     &     3.66  &         0.00        &        0.67    &        2.22     \\
1.0/2.0 &      -1.5  &        0.364     &    -4.12  &         0.00        &       -2.22    &       -0.79     \\
2.0/3.0 &      -1.373  &        0.393     &    -3.49  &         0.00        &       -2.14    &       -0.6     \\
3.0/4.0 &      -2.956  &        0.991     &    -2.98  &         0.00        &       -4.9    &       -1.01     \\
4.0/5.0 &      -2.147  &        0.694     &    -3.09  &         0.00        &       -3.51    &       -0.79     \\
\bottomrule
\end{tabular}}

\vspace{0.5em}

We used an analogous model to test the influence of education on estimated risk. The obtained log-likelihood was $-210.71$, the AIC $433.4$ and the BIC $451$. The model had 133 residuals and 6 degrees of freedom.

\resizebox{0.49\textwidth}{!}{%
\begin{tabular}{lcccccc}
\toprule
                  & coef & std err & z & P$> |$z$|$ & [0.025 & 0.975]  \\
\midrule
Education &      0.046  &        0.057     &     0.81  &         0.42        &       -0.07    &        0.16     \\
-1.0/1.0 &      -2.023  &        0.342     &    -5.91  &         0.00        &       -2.69    &       -1.35     \\
1.0/2.0  &       0.321  &        0.199     &     1.61  &         0.11        &       -0.07    &        0.71     \\
2.0/3.0  &      -0.175  &        0.135     &    -1.29  &         0.2        &       -0.44    &        0.09     \\
3.0/4.0  &      -0.086  &        0.132     &    -0.65  &         0.51        &       -0.34    &        0.17     \\
4.0/5.0  &      -0.452  &        0.234     &    -1.93  &         0.05        &       -0.91    &        0.01     \\
\bottomrule
\end{tabular}}
 
 \vspace{2em}
 \myparagraph{Gender} In this section, we divided the sample into female (left) and male (right) participants.
 
 \begin{tabular}{lrrrrr} \\ 
 \toprule 
  \multicolumn{2}{l}{Female} & \multicolumn{2}{l}{Male} \\
 \cmidrule(r){1-2} \cmidrule(r){3-4}
 $\mu$ ($\pm $sdt.) & \# &  $\mu$ ($\pm $sdt.) & \# & U & $p$ \\
 \midrule 
\multicolumn{4}{l}{Estimated Risk (\question{3})} \\ 
2.7 ($\pm$ 0.95) & 20 & 2.46 ($\pm$ 1.26) & 99 & 1098 & 0.43 \\ 
 \addlinespace \multicolumn{4}{l}{exposure (\question{2})} \\ 
0.4 ($\pm$ 1.16) & 20 & 0.6 ($\pm$ 1.43) & 99 & 945 & 0.64 \\ 
 \addlinespace \multicolumn{4}{l}{Poisoning Rating (\question{5})} \\ 
2.1 ($\pm$ 1.34) & 20 & 2.48 ($\pm$ 1.24) & 99 & 846 & 0.28 \\ 
 \addlinespace \multicolumn{4}{l}{Evasion Rating (\question{7})} \\ 
1.65 ($\pm$ 1.53) & 20 & 2.11 ($\pm$ 1.59) & 99 & 811 & 0.19 \\ 
 \addlinespace \multicolumn{4}{l}{Backdoor Rating (\question{9})} \\ 
0.55 ($\pm$ 1.63) & 20 & 1.99 ($\pm$ 1.62) & 99 & 517 & \textbf{0.0} \\ 
 \addlinespace \multicolumn{4}{l}{Sanity Rating (\question{11})} \\ 
-0.15 ($\pm$ 1.53) & 20 & 0.31 ($\pm$ 1.91) & 99 & 867 & 0.37 \\ 
 \addlinespace \multicolumn{4}{l}{Membership Rating (\question{13})} \\ 
1.25 ($\pm$ 1.76) & 20 & 1.83 ($\pm$ 1.79) & 99 & 793 & 0.16 \\ 
 \addlinespace \multicolumn{4}{l}{M. Stealing Rating (\question{15})} \\ 
0.85 ($\pm$ 1.96) & 20 & 1.94 ($\pm$ 1.72) & 99 & 673 & \textbf{0.02} \\ 
 \bottomrule
 \end{tabular}

\end{document}